\documentclass[manuscript]{acmart}
\setcopyright{none}

\usepackage{tabularx}
\usepackage{booktabs}
\usepackage{adjustbox}
\usepackage{enumitem}
\usepackage{subcaption}
\usepackage{multirow}
\usepackage{caption}
\usepackage{float}
\usepackage[ruled,vlined]{algorithm2e}

\title{Affect and Effect: Limitations of Regularisation-Based Continual Learning in EEG-based Emotion Classification}

\author{Nina Peire}
\affiliation{
  \institution{Imperial College London}
  \department{Department of Computing}
  \city{London}
  \country{United Kingdom}
}
\email{nina.peire24@imperial.ac.uk}

\author{Yupei Li}
\affiliation{
  \institution{Imperial College London}
  \department{Department of Computing}
  \city{London}
  \country{United Kingdom}
}
\email{yupei.li22@imperial.ac.uk}

\author{Björn Schuller}
\affiliation{
  \institution{Imperial College London}
  \department{Department of Computing}
  \city{London}
  \country{United Kingdom}
}
\email{bjoern.schuller@imperial.ac.uk}

\begin{document}

\begin{abstract}
Generalisation to unseen subjects in EEG-based emotion classification remains a challenge due to high inter-and intra-subject variability. Continual learning (CL) poses a promising solution by learning from a sequence of tasks while mitigating catastrophic forgetting. Regularisation-based CL approaches, such as Elastic Weight Consolidation (EWC), Synaptic Intelligence (SI), and Memory Aware Synapses (MAS), are commonly used as baselines in EEG-based CL studies, yet their suitability for this problem remains underexplored. This study theoretically and empirically finds that regularisation-based CL methods show limited performance for EEG-based emotion classification on the DREAMER and SEED datasets. We identify a fundamental misalignment in the stability-plasticity trade-off, where regularisation-based methods prioritise mitigating catastrophic forgetting (backward transfer) over adapting to new subjects (forward transfer). We investigate this limitation under subject-incremental sequences and observe that: (1) the heuristics for estimating parameter importance become less reliable under noisy data and covariate shift, (2) gradients on parameters deemed important by these heuristics often interfere with gradient updates required for new subjects, moving optimisation away from the minimum, (3) importance values accumulated across tasks over-constrain the model, and (4) performance is sensitive to subject order. Forward transfer showed no statistically significant improvement over sequential fine-tuning ($p > 0.05$ across approaches and datasets). The high variability of EEG signals means past subjects provide limited value to future subjects. Regularisation-based continual learning approaches are therefore limited for robust generalisation to unseen subjects in EEG-based emotion classification.

Code can be found at: https://github.com/glam-imperial/AffectEffect
\end{abstract}


\maketitle

\section{Introduction}
\label{sec:introduction}

Emotions shape our perceived reality and impact our every experience, making their study central to understanding the human condition. Emotion-aware systems support the study of emotions, by using models to label human emotional states from input data such as physiological signals. Their application extends into clinical tools \cite{Kamble2023} and 
brain-computer-interface (BCI) tasks such as motor imagery, which share similar EEG feature representations and modelling strategies \cite{lawhern_eegnet_2018, zhou_brainuicl_2024, duan_retain_2024}.

While emotion classification has seen significant progress over the past decade, the ability of models to generalise to unseen subjects remains a challenge \cite{li_personalized_2025, zhang_mini_2024, zhou_brainuicl_2024}. Due to the inherent physiological differences between individuals, as well as changes within a person over time, EEG signals exhibit covariate shift \cite{saha_intra-_2020}, characterised by a change in data distribution across and within subjects. When a machine learning model is sequentially fine-tuned on such data, it adapts to each new distribution, forgetting the previously acquired representations \cite{li_personalized_2025}, a phenomenon referred to as catastrophic forgetting. Continual Learning (CL) addresses this problem by introducing training strategies to preserve representations from previous subjects while enabling adaptation to new ones.

There exist various CL approaches: regularisation-based, replay-based, optimisation-based, representation-based, and architecture-based \cite{wang_comprehensive_2024} ones.
This study focuses on regularisation-based approaches, which penalises changes to parameters that were important to previous tasks \cite{van_de_ven_three_2022}. While replay-based approaches are the most common among recently-published EEG-based emotion classification studies \cite{li_personalized_2025}, they are more privacy-sensitive and require large memory buffers \cite{li_personalized_2025, kann_evaluation_2023}. With stricter regulations on data usage of AI systems \cite{EU_AI_act}, methods that store less sensitive data are highly relevant. Regularisation-based approaches are conceptually appealing due to their simplicity, no need for sample storage across tasks, nor network expansion \cite{kann_evaluation_2023}; and are often used in baseline comparisons, but their effectiveness for EEG-based emotion classification remains largely unexplored.

Elastic Weight Consolidation (EWC), Synaptic Intelligence (SI), and Memory Aware Synapses (MAS) are commonly used regularisation-based approaches, where each method introduces a different importance heuristic. By penalising changes to important parameters while allowing others to remain flexible, these methods directly address the stability–plasticity dilemma of retaining past representations while learning new tasks.

While CL is highly promising for EEG emotion classification \cite{li_continual_2024}, this remains largely uncharted territory, with several studies within the past year claiming ``little exploration has been made on [continual learning's] effectiveness to tackle subject shifts for clinical prediction tasks" \cite{duan_retain_2024}, and ``very few studies have been conducted in the literature on [Online Continual Learning], particularly in the context of EEG data"\cite{ahmad_robust_2025}. Most studies including subject-wise CL baselines use EWC in their baseline comparison \cite{zhou_brainuicl_2024, ahmad_robust_2025, duan_online_2024, aguilar_continuous_2025}, and 
\cite{cognemosense} implements EWC with modifications specifically for EEG-based emotion classification. However, to the best of our knowledge, no study has investigated the ability of regularisation-based CL methods to balance the stability-plasticity dilemma, by evaluating both backward and forward performance, in this domain.

Preliminary experiments in this study on the DREAMER and SEED datasets determined that EWC, SI, and MAS perform suboptimally for EEG-based emotion classification (Section~\ref{sec:preliminary}). Their performance is measured using average accuracy (ACC), the average `performance'
on current and past subjects, backward transfer (BWT), the change in performance of past tasks with respect to their performance upon learning, and forward transfer (FWT), the performance on unseen subjects relative to random initialisation. Not one method showed statistically significant improvements from the sequential fine-tuning baseline on FWT; and although BWT improved, current subject performance substantially degraded, a clear sign of the stability-plasticity dilemma. ACC showed minimal gains, primarily a result of its bias towards the increased BWT. These findings suggest that, despite their popularity as baselines in the literature, regularisation-based approaches are fundamentally limited for forward performance and scalability across large subject streams.

This study therefore investigates these limitations of regularisation-based continual learning for EEG-based emotion classification both theoretically and empirically (Sections~\ref{sec:3theoretical}–\ref{sec:results}).

\section{Related Work}
\label{sec:related-work}

Deep Learning has been instrumental in identifying underlying patterns in EEG signals for emotion classification. In 2015, Zheng et al.\ \cite{zheng2015investigating} were among the first to use a deep learning architecture for EEG-based emotion classification, proving its ability to outperform an SVM. EEGNet \cite{lawhern_eegnet_2018} was later introduced as a unified Convolutional Neural Network architecture designed to effectively classify EEG signals across various BCI paradigms. Using raw input data, EEGNet extracts neurophysiologically interpretable features through spatial and temporal convolutional filters.
To capture continuous long-range dependencies in both spatial and temporal features, recurrent models such as RNNs, LSTMs, and transformers have since been used in EEG-based emotion classification tasks \cite{tao_eeg-based_2023, hu_eeg-based_2022}. There exist many similar studies, exploring different deep learning approaches to tackling EEG-based emotion classification \cite{roy_deep_2019}.
Much of this research has focused on outperforming baselines within specific datasets, often by training and testing on the same dataset, or using a subject-dependent approach, rather than developing tools which can generalise across different people and real-world conditions.
This generalisability in a subject-independent approach is a major challenge to EEG-based emotion classification due to the high inter- and intra-subject variability.

This is an open issue in the literature which has sparked significant research into how to mitigate this, as generalisation to unseen subjects is key to practical deployment \cite{roy_deep_2019}. EEG signals are inherently non-stationary, with variability across time, spectral channels, and individuals due to factors such as age, physiology, and testing environment \cite{chaddad_electroencephalography_2023, apicella_toward_2024, Smith6136}. The variability causes distribution shifts between training and test data \cite{saha_intra-_2020, krusienski_critical_2011}, making subject-independent classification significantly harder than subject-dependent classification.
Recent works have identified the potential of Continual Learning (CL) for adapting EEG-based emotion classification to real-world systems. 
Duan et al.\ \cite{duan_online_2024, duan_retain_2024} claim being first to tackle the challenge of online sequential decoding of EEG classification. Their MUDVI model combines a constrained memory buffer with kernel based subject shift detection. Their second model AMBM, uses a meta-learning strategy with an adaptive meta optimiser and bi-level mutual information (MI) maximisation. AMBM outperformed other state-of-the-art continual learning methods such as A-Gem \cite{chaudhry_efficient_2019} achieving 12.37\% and 20.33\% higher accuracy relative to sequential fine-tuning on DEAP and SEED data respectively, on 4-class (arousal-valence model quadrants) 
and 3-class (negative, neutral, positive) classification, respectively \cite{duan_retain_2024}. They do, however,  point out that these continual learning methods are not yet able to meet the upper baseline of joint training. Additionally, Duan et al.\ \cite{duan_online_2024} further reported that EWC outperforms sequential fine-tuning on both datasets. However, they only report train-test performance on the same set of subjects and lack any results to unseen subjects or forwards transfer. Li et al.\ \cite{li_personalized_2025} explicitly state the potential of CL methods to reduce catastrophic forgetting caused by inter-subject variability in EEG signals. More recently Zhou et al.\ \cite{zhou_brainuicl_2024} proposed BrainUICL, an Unsupervised Individual Continual Learning paradigm, which aims to continuously adapt to incoming new subjects. Their proposed architecture consists of a Dynamic Confident Buffer (DCB) and Cross Epoch Alignment (CEA). While their method is effective across BCI paradigms, their evaluation showed much lower performance for emotion classification, underscoring its difficulty. Finally, a recent preprint \cite{cognemosense} relies on the use of EWC for EEG-based emotion classification, to retain knowledge acquired from previous subjects while adapting to new subjects with minimal retraining. 
The work shows the potential of exploring regularisation-based CL approaches for EEG-based emotion classification.

Regularisation-based CL approaches protect previously acquired knowledge by adding a quadratic penalty to the loss function. This is formalised by:
\begin{equation*}
    \label{eq: regularisation}
    \widetilde{\mathcal{L}}_\tau(\theta) = \mathcal{L} _\tau(\theta) + \lambda\sum_k^{N_{\text{params}}} \Omega_k^{(\tau-1)} \left( \theta_k - \theta_k^{*(\tau-1)} \right)^2
\end{equation*}
for a given task $\tau$, where $\lambda>0$ is the regularisation-strength hyperparameter, $\mathcal{L}_\tau(\theta)$ is the standard loss (e.g., cross entropy), and $\theta_k^{*(\tau-1)}$ is the optimal parameter $k$ of the previous task. Typically, $\Omega_k^{\tau} = \Omega_k^{(\tau-1)}+\Omega_k$, where $\Omega_k^{(\tau-1)}$ is the cumulative importance of parameter $k$ from all previously learnt tasks, and $\Omega_k$ is the importance contribution from the current task $\tau$ \cite{benzing_unifying_2022}. EWC \cite{kirkpatrick_overcoming_2017}, SI \cite{zenke_continual_2017}, and MAS \cite{aljundi_memory_2018} each follow this framework, with unique importance heuristics for $\Omega_k$.

EWC uses the Fisher Information Matrix (FIM) (Eq.~\eqref{eq:fim}) as a heuristic for parameter importance, because under regularity conditions, the Fisher Information equals the expected Hessian of the negative log-likelihood (NLL) \cite{pascanu_revisiting_2014, kunstner_limitations_2019}.
\begin{equation}
    F(\theta) = \frac{1}{n}\sum^n_{i=1}\mathbb{E}_{p_\theta(y|x_i)}\left[\nabla_\theta \log p_\theta (y|x_i)\nabla_\theta \log p_\theta (y|x_i)^\top\right]
    \label{eq:fim}
\end{equation}
The Hessian matrix of the negative log-likelihood is a direct measure of local curvature of the loss landscape \cite{meshkov_convnets_2024}. This is formally defined, per \cite{pmlr-v108-barshan20a, kirkpatrick_overcoming_2017, pascanu_revisiting_2014, kunstner_limitations_2019} in the Appendix \ref{sec:fisher-hessian}.
Parameters with high $F_{kk}$ are considered highly important for the previously learnt task, as small deviations in their values lead to large increases in loss. EWC penalises changes to these important parameters, thereby preserving task-specific knowledge when learning subsequent tasks.
As noted by Van De Ven \cite{ven_computation_2025}, most studies rarely describe how they compute the Fisher Information, despite multiple implementation variants existing. He points out that EWC is commonly used as a baseline even though the original paper lacked a fully detailed implementation. 

SI uses the following importance heuristic:
\begin{align*}
    \Omega_k^\tau = \sum_{\nu < \tau} \frac{\omega_k^\nu}{(\Delta_k^\nu)^2 + \xi}, \quad\quad
    \omega^\nu_k&\approx -\sum_t g_{k,t}\Delta\theta_k = -\sum_t  \frac{\partial L}{\partial \theta_k}(\theta_{k, t+1}-\theta_{k,t}), \quad\quad
    \Delta_k^\nu \equiv \theta_k(t^\nu) - \theta_k(t^{\nu-1}), 
    \label{eq:si-importance}
\end{align*}
where $\omega_k^{\nu}$ is the path integral and $\Delta_k^\nu$ the parameter change across task $\nu$. Parameters that have large gradient magnitudes are therefore given higher importance, whereas those with little impact on the loss are given low importance.

MAS accumulates importance for each parameter based on how sensitive the network's output is to a small change in each parameter \cite{aljundi_memory_2018}.
The surrogate loss is computed without requiring labelled data, contrary to EWC and SI. Weight importance is defined by:
\begin{equation}
\label{eq:avg}
    \Omega_k = \frac{1}{n} \sum_{i=1}^{n} \| g_k(x_i) \|, 
\end{equation}
where $n$ is the total number of data points at a given phase and $g_k(x_i)$ is the gradient of the learnt function with respect to the parameter $\theta_k$ for input $x_i$ \cite{aljundi_memory_2018}. Since this is computationally very inefficient, the authors suggest using the gradients of the squared $\ell_2$-norm of the learnt function output: 
\begin{equation}
\label{eq:mas_grad}
    g_k(x_i) = \frac{\partial [\ell^2_2(F(x_i;\theta))]}{\partial \theta_k}.
\end{equation}
A large $\Omega_k$ indicates that small perturbations to $\theta_k$ strongly affect the network output, irrespective of its impact on the loss function.

Benzing \cite{benzing_unifying_2022} conducted a thorough theoretical and empirical analysis of exactly these techniques (EWC, SI, and MAS), demonstrating that despite their differing importance heuristics, they are each related to the Fisher Information and depend on a bias derived from gradient variance. Zhao et al.\  \cite{zhao_statistical_2024} further demonstrated that under imperfect conditions like domain shifts, data heterogeneity, or high noise, regularisation-based continual learning methods face a larger trade-off between stability (retaining previous knowledge) and plasticity (learning new knowledge).

While increasingly many studies employ regularisation-based approaches, there has not been any investigation into their effectiveness for EEG-based emotion classification.

\section{Theoretical Grounding for Sub-optimal Performance}
\label{sec:3theoretical}
We now aim to explain and support the limitations of regularisation-based approaches under high inter-and intra-task variability, from a mathematical underpinning. 

This study defines inter-task variability as the covariate shift between subjects, while intra-task variability is defined by EEG signal noise (e.g., intra-subject variability, physiological artefacts, electrode impedance, and skull attenuation)---which interacts with stochastic noise inherent to the optimisation process.

\subsection{Effects of intra-task variability and stochastic noise on importance heuristic}

\vspace{2mm}\par\noindent\textbf{Elastic Weight Consolidation:}
\label{sec:ewc}
\label{sec:fisher}
Since computing the true FIM is intractable, EWC typically relies on empirical Fisher Information ($\widetilde{F}$) \cite{ven_computation_2025}, as did experiments in this study. The empirical Fisher Information is defined as \cite{kunstner_limitations_2019}:
\begin{equation}
    \widetilde{F}(\theta) := \frac{1}{n}\sum_{i=1}^n \nabla_{\theta} \log p_{\theta}(y_i | x_i) \nabla_{\theta} \log p_{\theta}(y_i | x_i)^\top
    .\label{eq:EF}
\end{equation}
As stressed by \cite{pascanu_revisiting_2014, kunstner_limitations_2019, ven_computation_2025}, $\widetilde{F}$ does not reliably approximate the traditional Fisher Information at the minimum.

By the law of large numbers, the dataset must be sufficiently large for $\widetilde{F}(\theta) \xrightarrow{N\rightarrow\infty}F(\theta)$. When the empirical Fisher Information uses a small number of samples to approximate it, the estimate is dominated by stochastic noise, which is worsened by intra-subject variability of EEG data. The decomposition of $\widetilde{F}_{kk}$ makes this explicit:
\begin{align}
    \widetilde{F}_{kk}(\theta) = \mathbb{E}[g^2_k] \equiv \text{Var}(g_k) + \mathbb{E}[g_k]^2,
    \label{eq:decomp}
\end{align}
where $g_k$ is the gradient of the NLL loss of the $k$th diagonal element. Thus, a large $\widetilde{F}_{kk}$ can result from either a consistently strong signal ($\mathbb{E}[g_k]$) or high gradient variance ($\text{Var}(g_k)$). This echoes Kunstner's claim \cite{kunstner_limitations_2019} that the empirical Fisher Information adapts to stochastic gradient noise rather than second-order curvature. Since $\widetilde{F}$ is computed with batch size 1 to allow gradients of individual data points to be squared per Equation~\eqref{eq:EF}, $\widetilde{F}$ suffers from the maximum stochastic gradient noise, inflating the gradient variance.
In noisy EEG data, this can inflate the estimated importance of parameters sensitive to noise, potentially drowning out true important parameters.

Additionally, the Fisher Information itself is a poor proxy for the expected Hessian matrix of the negative log-likelihood (true curvature), under model misspecification \cite{kunstner_limitations_2019}. Kunstner affirms that $\widetilde{F}$ or $F$ only approaches the Hessian matrix under the following rarely satisfied conditions: the likelihood is well-specified, the model is capable of representing the data, the prediction function captures all relevant information, and the model has converged to a minimum of the NLL, where $p_\theta(y|x)\approx p_\text{true}(y|x)$ \cite{kunstner_limitations_2019}. Under highly dynamic, low-SNR, and non-i.i.d.\ data like EEG signals, these conditions are particularly implausible. In EWC, this could lead to excessive regularisation of parameters sensitive to noise rather than meaningful features.

\vspace{2mm}\par\noindent\textbf{Synaptic Intelligence:}
\label{sec:si}
Since the path integral is computed at step-level rather than task boundaries, stochastic noise from batched gradient descent and inherent intra-subject variability from EEG signals directly impacts the importance estimate. Under SGD, the path integral becomes:
\begin{align}
    \omega^\nu&\approx \eta\sum_{t\leqslant T} (g_t + \sigma_t + \mu_t)^2 
    =\eta \sum_{t \leqslant T} \left( g_t^2 + \underbrace{\sigma_t^2 + \mu_t^2 + 2g_t\sigma_t + 2g_t\mu_t + 2\sigma_t\mu_t}_{\text{Error introduced by noise}} \right),
\end{align}
where $\sigma_t$ is zero-mean stochastic noise and $\mu_t$ EEG signal noise. Since the squared noise terms are always positive, the path integral may inflate due to increased gradient variance caused by noise rather than meaningful signal.

As derived in Appendix~\ref{sec:path-integral-adam-math}, the path integral under Adam similarly depends on the accumulation of squared gradients \cite{benzing_unifying_2022}:
\begin{align}
\label{eq:si_decomp}
    \omega^\nu&\approx \sum_{t\leqslant T}\frac{\eta_t(1-\beta_1)(g_t + \sigma_t + \mu_t)^2}{\sqrt{v_t}}
    = \frac{1-\beta_1}{\sqrt{v_T}}\sum_{t\leqslant T}\eta_t\sqrt{\frac{v_T}{v_t}}(g_t + \sigma_t + \mu_t)^2.
\end{align}
While $\sqrt{v_t}$ acts as a normaliser, being the exponential moving average of squared gradients \cite{kingma_adam_2017}, it adapts with a delay determined by $\beta_2$. Hence, sudden spikes in noisy gradients immediately inflate the numerator while the denominator's response is smoothed out across time. As a result, the path integral may accumulate with gradient variance and show erratic fluctuations due to artefacts.

\vspace{2mm}\par\noindent\textbf{Memory Aware Synapses:}
\label{sec:mas}
MAS is more robust to stochastic noise.
By using gradients from the norm (Eq.~\eqref{eq:mas_grad}) and then computing their absolute value, MAS avoids the quadratic amplification of gradient noise present in EWC and SI. Averaging these absolute gradients over tasks (Eq.~\eqref{eq:avg}) further stabilises the importance estimates according to the law of large numbers.

However, MAS does face a label and input noise trade-off.
Sample labels in EEG-based emotion classification are typically self-reported, making them prone to label noise. Since the MAS surrogate loss is computed without labels, explicit label noise is omitted. This also avoids complications caused by loss being in a local minimum where gradients are near zero and parameter updates stall \cite{aljundi_memory_2018}. However, this increases reliance on the input data, which is inherently noisy, too. Under model misspecification or label noise, the confidence of outputs may be weak, which will impact gradient computation and therefore the importance estimates. In EEG signals with a low-SNR, artefacts such as muscle spikes, and heartbeats can produce large output changes for small parameter perturbations, artificially inflating $\Omega_k$. Therefore, while MAS is more robust to stochastic noise, it is more sensitive to outlier samples.

\subsection{Effects of inter-task variability on plasticity and generalisability}
High inter-subject variability increases the likelihood that different subjects have more distant parameter optima. After training subject A, parameter $\theta_k$ may be consolidated at its value $\theta^*_{A, k}$. If the optimum for subject B, $\theta^*_{B,k}$, is considerably different, it will already be constrained by the quadratic penalty, limiting the model's plasticity. This occurs because the optimisation objective itself is changed from minimising $\mathcal{L}_B(\theta)$ to minimising $\widetilde{\mathcal{L}}_B(\theta)$:
\begin{align}
    \nabla\widetilde{\mathcal{L}}_B(\theta) &= \nabla\mathcal{L}_B(\theta) + \lambda\boldsymbol{\Omega}(\theta - \theta^*_A) = 0
    \Rightarrow \nabla\mathcal{L}_B(\theta) = - \lambda\boldsymbol{\Omega}(\theta - \theta^*_A).
\end{align}
This optimum is a compromise where the regularisation-based approach tries to minimise the loss for task B while staying near the optimal parameters of task A. Therefore, $\nabla\mathcal{L}_B(\theta)=0$ is never reached unless $\lambda = 0$ or $\theta^*_B=\theta^*_A$, which is highly unlikely under high inter-subject variability. Additionally, a larger distance between optima by definition increases the magnitude of the quadratic penalty. This strengthens the regularisation penalty which further reduces plasticity, additionally biasing the model towards stability for the previous task at the cost of adapting to the incoming one, limiting generalisability in the process.

\subsection{Effects of importance accumulation on scalability}
The importance $\Omega_k$ is never reset and only accumulates across tasks. As the number of tasks increases, the parameters become progressively more constrained over time, essentially ``freezing" them and making plasticity negligible. This makes scalability in synaptic intelligence inherently limited for large streams of tasks.

While Online EWC \cite{huszar_note_2018} avoids the unbounded linear growth of the direct sum $F_t^* = F_t + F_{t-1}$ by replacing this with a running update: $F^*_t = \gamma F^*_{t-1}+F_t$ \cite{schwarz_progress_2018}, importance from previous tasks persist in the running average, weighted by $\gamma$. This accumulation can still over-constrain the network, or cause parameters to freeze.

\section{Hypotheses}
\label{sec:hypotheses}
Based on the theoretical grounding in Section~\ref{sec:3theoretical} and preliminary experimental findings in Section~\ref{sec:preliminary} which reveal limited performance of EWC, SI, and MAS consistent with the stability-plasticity dilemma, we form the following hypotheses on the limitations of regularisation-based CL approaches for EEG-based emotion classification.
\begin{itemize}[left = 0pt]
    \item \textbf{H1:} The importance heuristics for each approach are less reliable and can inflate under EEG signal noise.
        \begin{itemize}
            \item \emph{(EWC)} Empirical Fisher Information deviates from the Fisher Information under smaller sample sizes; and the Fisher Information deviates from the expected Hessian under model misspecification.
            \item \emph{(SI)} The path integral inflates under stochastic noise and EEG signal noise (incl.\ intra-subject variability).
            \item \emph{(MAS)} Importance estimates are more robust under stochastic noise.
        \end{itemize}
    \item \textbf{H2:} Subsequent subjects require conflicting parameter updates, leading to gradient interference rather than convergence, limiting generalisability.
    \item \textbf{H3:} Importance accumulates across tasks, constraining plasticity in the network early, limiting scalability.
    \item \textbf{H4:} Subject order impacts which parameters are constrained early, making each method highly sensitive to random seed.
\end{itemize}

\section{Method}
\label{sec:method}


This study implements the \textbf{Online EWC} variant \cite{kirkpatrick_overcoming_2017, huszar_note_2018, schwarz_progress_2018}, which updates a running Fisher Information Matrix (FIM) after each task. The FIM in this study is estimated by the empirical Fisher Information which is computed at task boundaries by calculating the squared gradients of the negative log-likelihood loss on a randomly sampled subset of 500 samples of the training data. The running FIM is then updated using this estimate according to $F^*_t = \gamma F^*_{t-1}+F_t$, with $\gamma=0.9$. During training of a new task, the running FIM is used to compute the penalty at every step, by multiplying the FIM value for each parameter with the squared distance between the current parameter value and the parameter at the end of training the previous task. This penalty is scaled by $\lambda$ and added to the loss function.

\textbf{SI} is implemented according to Zenke et al.\ \cite{zenke_continual_2017}, which measures parameter importance by accumulating a path integral during training, for each task. This is computed by summing the product of the gradient and parameter update at every training step. This value is used to compute the task importance at the end of the task by normalising the final path integral with the squared change in the parameter's value between the start and end of the task, as well as a small dampening hyperparameter to prevent division by zero. The total importance for each parameter is the sum of its importance values from all previous tasks. This importance estimate is used to compute the penalty and total loss in the same manner as EWC.

\textbf{MAS} follows Aljundi et al.\ \cite{aljundi_memory_2018}, and measures parameter importance at the end of every task by iterating over the entire training dataset and computing the mean absolute gradients of the norm of the model output logits with respect to each parameter. The importance is accumulated by summing the new importance values with those from previous tasks. This total importance estimate is then used to compute the penalty and total loss in the same manner as EWC and SI.

An additional \textbf{Naïve} strategy is implemented which adds no regularisation.

\section{Experiment Design}
\label{sec:experimental-design}

\subsection{Datasets}
\label{sec: dataset_descriptions}
This study uses the publicly available DREAMER \cite{dreamer} and SEED \cite{zheng2015investigating, duan2013differential} datasets.

The DREAMER dataset \cite{dreamer} consists of EEG recordings from 23 subjects (14 male, 9 female) between ages 22-33 years (M = 26.6 years, SD = 2.7 years). Each subject watched 18 movie clips to evoke emotional responses, during a single recording session. The signals were recorded using a portable, wearable, wireless, low-cost headset, making the results more representative of future publicly available wearable devices. The participants reported their emotions using self-assessed valence, arousal, and dominance scores on a scale of 1 to 5.
The EEG signals were measured at a 128\,Hz sampling rate across 14 channels according to the International 10-20 system.

The SEED Dataset \cite{zheng2015investigating, duan2013differential} contains EEG signals from 15 subjects (7 male, 8 female; M = 23.27 years, SD = 2.37 years. Participants watched 6 clips, repeated across 3 sessions. The signals were collected using the 62-channel ESI NeuroScan System at 1\,kHz, then downsampled to 200\,Hz. The signals were labelled through self-assessment following the categories: negative, neutral, or positive.

The SEED dataset contains signals of much higher temporal and spatial resolution than the DREAMER dataset, due to its higher channel count and sampling frequency. This results in the SEED dataset being able to capture more detailed neural activity than the DREAMER dataset. Conversely, DREAMER’s use of a portable, low-density headset makes it more representative of wearable BCI applications, though potentially less capable of capturing fine-grained patterns.

\subsection{Data Pre-processing}
This study's pre-processing pipelines for DREAMER and SEED are outlined in Algorithms~\ref{alg:pre-processing-dreamer},\ref{alg:pre-processing-seed}.

\subsubsection{DREAMER}
Artefact removal is essential to EEG pre-processing due to the high signal-to-noise ratio where noise often has similar amplitudes to neural activity \cite{roy_deep_2019}. There exist several techniques to eliminate artefacts after separating components with ICA. This study uses the kurtosis value per component to identify outlier signals \cite{7391935}. The kurtosis of activity indicates peaks in the activity distribution \cite{DELORME20049}, which can be removed at a given threshold. It was observed that a z-scored kurtosis threshold of 4.0 was able to remove some components without removing too many. While this technique remains an available practice in EEGLab, it is more commonplace recently to use methods such as ICLabel or manual removal instead. However, both of these approaches require expert knowledge on EEG signals, a limitation further described in Section~\ref{sec:limitations}. The cleaned signals are then segmented into 3\,s windows with 1\,s overlap.
In line with subject-incremental continual learning, the data is partitioned at the subject level. Subjects are randomly split 80:20 into a train-test stream and an unseen held-out set. Within each subject, samples are split 80:20 into train and test using the TorchEEG library function \verb+train_test_split_groupby_trial+. No separate development (validation) set was used, as all hyperparameters were fixed a priori in accordance with common continual learning practices and preliminary experiments. Models were trained for a fixed number of epochs without early stopping, ensuring that each subject contributed equally to parameter updates. All random splits and model initialisations were performed using fixed random seeds.

\begin{algorithm}[hbt!]
\footnotesize
\caption{DREAMER Pre-processing Pipeline}
\label{alg:pre-processing-dreamer}
\LinesNotNumbered
\SetAlgoLined
\KwIn{$X$: raw EEG signals (14 channels, 128\,Hz)}
\KwOut{$Z$: cleaned epochs; $y$: labels}
\BlankLine

\textbf{Build Independent Component Analysis dictionary: }\;
\ForEach{subject $s$}{
BaselineRemoval$(X)$; \\
NotchFilter$(X, fs{=}128, f_0{=}50, q = 30)$; \\
ButterworthFilter$(X, 0.5{-}45\,\text{Hz}, \text{order }4, fs{=}128)$; \\
concatenate all trials into a single signal vector $\to X_s$; \\
fit FastICA with $n_c=\text{channels}$, whiten=unit-variance; \\
$\text{ICA\_dict}[s] \leftarrow \text{fitted ICA}$\;
}
\BlankLine

\textbf{Data Cleaning: }\;
\ForEach{subject $s$, trial}{
BaselineRemoval$(X)$; \\
NotchFilter$(X, fs{=}128, f_0{=}50, q = 30)$; \\
ButterworthFilter$(X, 0.5{-}45\,\text{Hz}, \text{order }4, fs{=}128)$; \\
$S \leftarrow \text{ICA}[s].\text{transform}(X^\top)$; \\
compute kurtosis per component; $z$-score; zero components with $|z|>4.0$; \\
$X \leftarrow \text{ICA}[s]^{-1}(S)^\top$; \\
MeanStdNormalize; \\
segment to epochs: chunk\_size $=384$  (3\,s), overlap $=128$ (1\,s); \\
To2d; \\
ToTensor; \\
}
\BlankLine

\textbf{Label Transform: }
\ForEach{sample in dataset}{
$(\text{arousal},\text{valence}) \to \{\text{HAHV},\text{LAHV},\text{HALV},\text{LALV}\} \to y$\;
}
\Return{$Z, y$}
\end{algorithm}

\subsubsection{SEED}
SEED's dataset already underwent minimal pre-processing by the authors \cite{zheng2015investigating, duan2013differential} which includes downsampling from 1\,kHz to 200\,Hz as well as a 0-75Hz band-pass filter. The choice was made not to downsample SEED's sampling frequency to match DREAMER's as this would reduce SEED’s temporal resolution and degrade signal quality.
The SEED dataset was split using the same subject-incremental strategy as DREAMER. Samples are similarly split into 80:20 train and test using TorchEEG's \verb+train_test_split_groupby_trial+. 

\begin{algorithm}[hbt!]
\footnotesize
\caption{SEED Pre-processing Pipeline}
\label{alg:pre-processing-seed}
\LinesNotNumbered
\SetAlgoLined
\KwIn{$X$: raw EEG signals (62 channels, 200 Hz)}
\KwOut{$Z$: cleaned epochs; $y$: labels}
\BlankLine

\textbf{Build Independent Component Analysis dictionary: }\;
\ForEach{subject $s$}{
    NotchFilter$(X, fs{=}128, f_0{=}50, q = 30)$; \\
    ButterworthFilter$(X, 0.5{-}45\,\text{Hz}, \text{order }4, fs{=}128)$; \\
    concatenate all trials into a single signal vector $\to X_s$; \\
    fit FastICA with $n_c=\text{channels}$, whiten=unit-variance; \\
    $\text{ICA\_dict}[s] \leftarrow \text{fitted ICA}$\;
}
\BlankLine

\textbf{Data Cleaning: }\;
\ForEach{subject $s$, session file}{
  NotchFilter$(X, fs{=}200, f_0{=}50, q=30)$; \\
  ButterworthFilter$(X, 0.5{-}45\,\text{Hz}, \text{order }4, fs{=}200)$; \\
  $S \leftarrow \text{ICA}[s].\text{transform}(X^\top)$; \\
  compute kurtosis per component; $z$-score; zero components with $|z|>4.0$; \\
  $X \leftarrow \text{ICA}[s]^{-1}(S)^\top$; \\
  MeanStdNormalize per channel; \\
  segment to epochs: chunk\_size $=600$ (3\,s), overlap $=200$ (1\,s); \\
  To2d; \\
  ToTensor; \\
}
\BlankLine

\textbf{Label Transform: }\;
\ForEach{sample in dataset}{
  $\{-1, 0, 1\}$ $\rightarrow$ $\{0, 1, 2\}$\;
}
\Return{$Z, y$}
\end{algorithm}

\subsection{Model Architecture}
\begin{figure}[hbt!]
    \centering
    \includegraphics[width=0.9\linewidth]{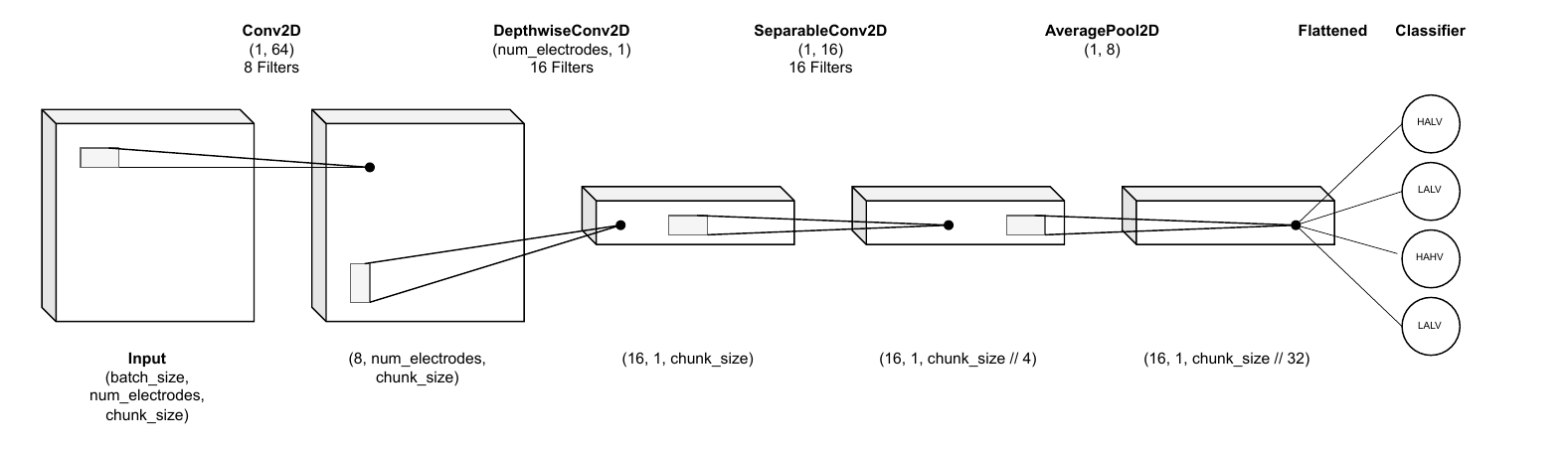}
    \caption{EEGNet Architecture Details as adapted from \cite{lawhern_eegnet_2018}}
    \label{fig:eegnet_both}
\end{figure}

This study adopts the existing CNN-based EEGNet-8,2 model because of its ability to extract neurophysiologically interpretable features through spatial and temporal convolutional filters, while maintaining a compact architecture. The architecture of EEGNet is outlined in Figure~\ref{fig:eegnet_both}. While many state-of-the-art studies employ transformer-based architectures, EEGNet's simple and compact architecture facilitates clearer interpretation of continual learning dynamics. While EEGNet may show lower performance \cite{bazargani_emotion_2023} than other CNN-based EEG models such as ShallowConvNet or DeepConvNet \cite{shallowconvnet}, its low parameter count provides a controlled setting to study the stability–plasticity trade-off.


\subsection{Evaluation Metrics}
\label{sec: metrics}
We evaluate each of the continual learning strategies using Average Accuracy (ACC), Backward Transfer (BWT), and Forward Transfer (FWT) \cite{lopez-paz_gradient_2022}. A train-test accuracy matrix, $R \in \mathbb{R}^{T \times T}$, is maintained where each entry $R_{i,j}$ is the test accuracy on the test set of task $t_j$ after completing training on task $t_1, ..., t_i$.
In addition to these metrics, this study will look at training accuracy and training F1-score as a proxy for plasticity as this is indicative of the ability of the model to change its weights.

\subsection{Technical Setup}
The Adam optimiser \cite{kingma_adam_2017} with learning rate 0.001 for training on DREAMER and 0.0002 for training on SEED with no weight decay was chosen for this study. Each model is trained for 30 epochs. The models are trained on a Tesla A30 24\,GB. The code is written in Python 3.12.3, with the ML architectures written using PyTorch 2.7.1 with CUDA 12.6 support, and using TorchEEG 1.1.2 for dataset and model handling.

\section{Results}
\label{sec:results}

\subsection{Preliminary Experiment Results}
\label{sec:preliminary}

Regularisation-strength ($\lambda$) tuning experiments yield the following key observations:

\begin{itemize}[left = 0pt]
    \item \textbf{Sequential fine-tuning learns a subject-dependent signal, which is catastrophically forgotten.} The train-test accuracy matrix in Figure~\ref{fig:naive_acc_matrix} explicitly shows the effect of catastrophic forgetting, where the current subject being learnt has high test-accuracy, but past or future subjects do not. Since the SEED dataset per subject contains samples from three recording sessions, it is evident it struggles to learn due to the higher intra-subject variability.
    \begin{figure}[hbt!]
        \centering
        \captionsetup{width=0.9\linewidth}
        \includegraphics[width=0.9\linewidth]{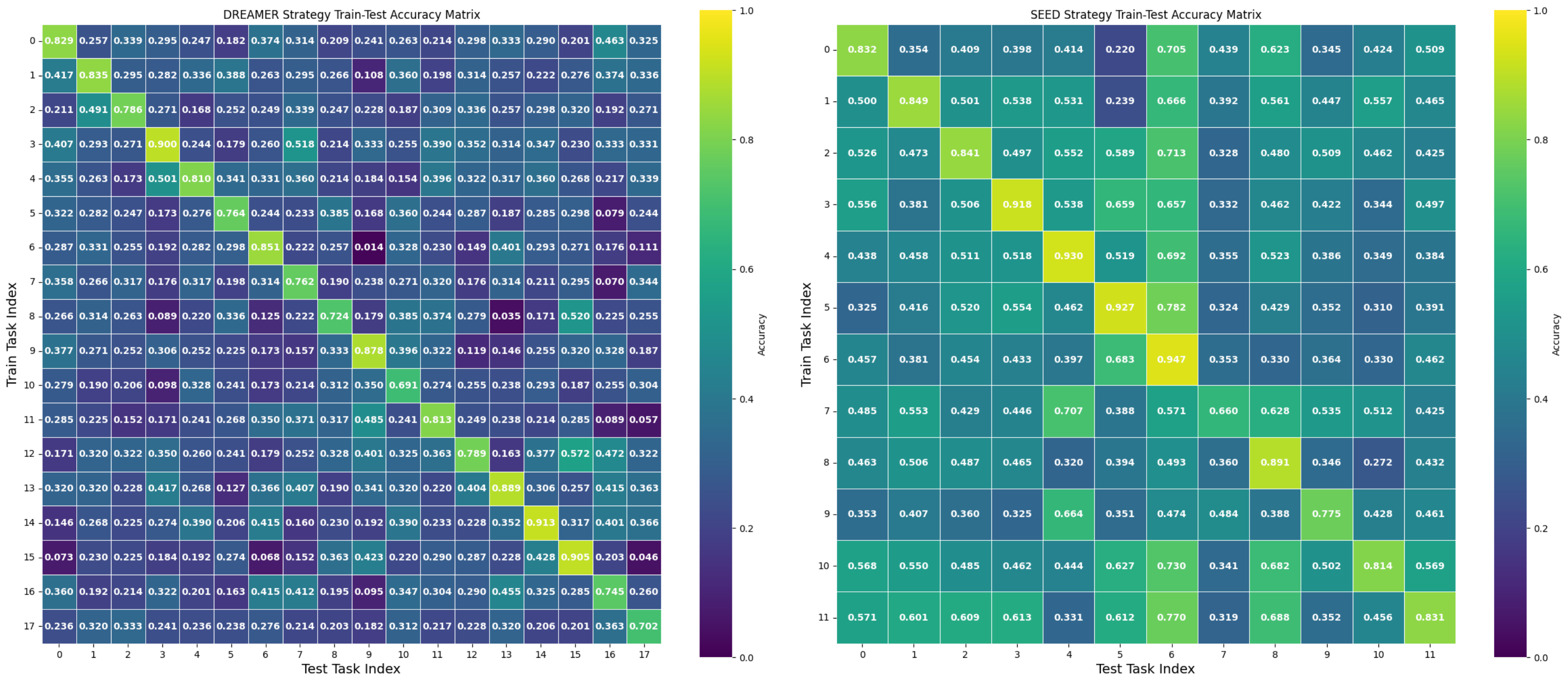}
        \caption{[Naïve] Train-test accuracy matrices of naïve subject-incremental learning, meaning without regularisation strategy, indicating the presence of catastrophic forgetting. The left plot shows results on the DREAMER dataset, and right plot those on the SEED dataset.}
        \label{fig:naive_acc_matrix}
    \end{figure}

    \item \textbf{Regularisation yields minimal improvements on average accuracy.} This is depicted in Figure~\ref{fig:dreamer_acc} for DREAMER and Figures~\ref{fig:seed_acc}, \ref{fig:seed_acc_lambda1} in the Appendix for SEED. Under little regularisation (small $\lambda$) average ACC across tasks approaches the baseline performance as expected. There is a peak where a certain amount of regularisation benefits average accuracy by improving backwards transfer (stability) without considerably harming current task performance (plasticity). Under stronger regularisation (large $\lambda$), performance drops as the network's ability to learn is too constrained. These gains, however, are near-negligible as is evident from Figure~\ref{fig:dreamer_acc_lambda1}, where overall ACC still performs similarly to the baseline.
    \begin{figure}[hbt!]
        \centering
        \captionsetup{width=0.85\linewidth}
        \includegraphics[width=0.85\linewidth]{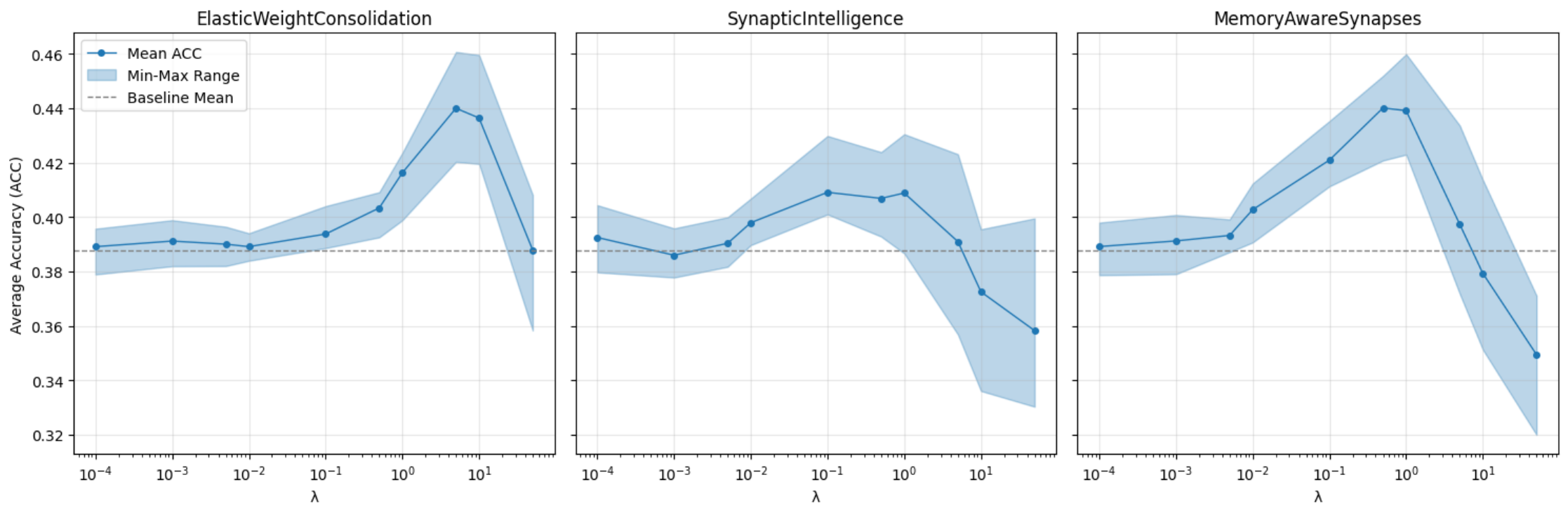}
        \caption{[DREAMER] Average ACC across five seeds for each strategy with respect to increased regularisation ($\lambda$) upon repeated training.}
        \label{fig:dreamer_acc}
    \end{figure}
    \begin{figure}[hbt!]
        \centering
        \captionsetup{width=0.85\linewidth}
        \includegraphics[width=0.85\linewidth]{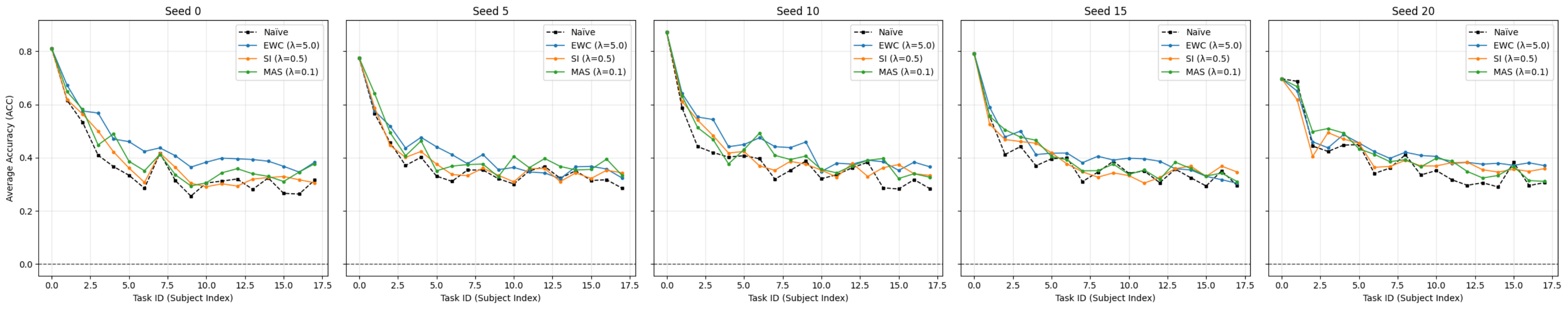}
        \caption{[DREAMER] ACC across tasks for each strategy and each seed, where EWC $\lambda = 5.0$, SI $\lambda = 0.5$, and MAS $\lambda = 0.1$.}
        \label{fig:dreamer_acc_lambda1}
    \end{figure}

    \vspace{2cm}
    \item \textbf{Backward Transfer increases with increased regularisation strength.} This is evident from Figure~\ref{fig:dreamer_bwt} for DREAMER and Figures~\ref{fig:seed_bwt}, \ref{fig:seed_bwt_lambda1}, \ref{fig:seed_f1} in the Appendix for SEED. This improved BWT is partially caused by the lower F1-scores across learning. Increased regularisation decreases the model's ability to learn (Figure~\ref{fig:dreamer_f1} in the Appendix) per the stability-plasticity dilemma, which inherently reduces model performance. When less is learnt, there is less to forget, such that BWT can see artificial gains (Figure~\ref{fig:dreamer_bwt_lambda1}). Each strategy shows improved BWT from the baseline ($p < 0.05$ for all strategies and datasets).

        \begin{figure}[hbt!]
            \centering
            \captionsetup{width=0.85\linewidth}
            \includegraphics[width=0.85\linewidth]{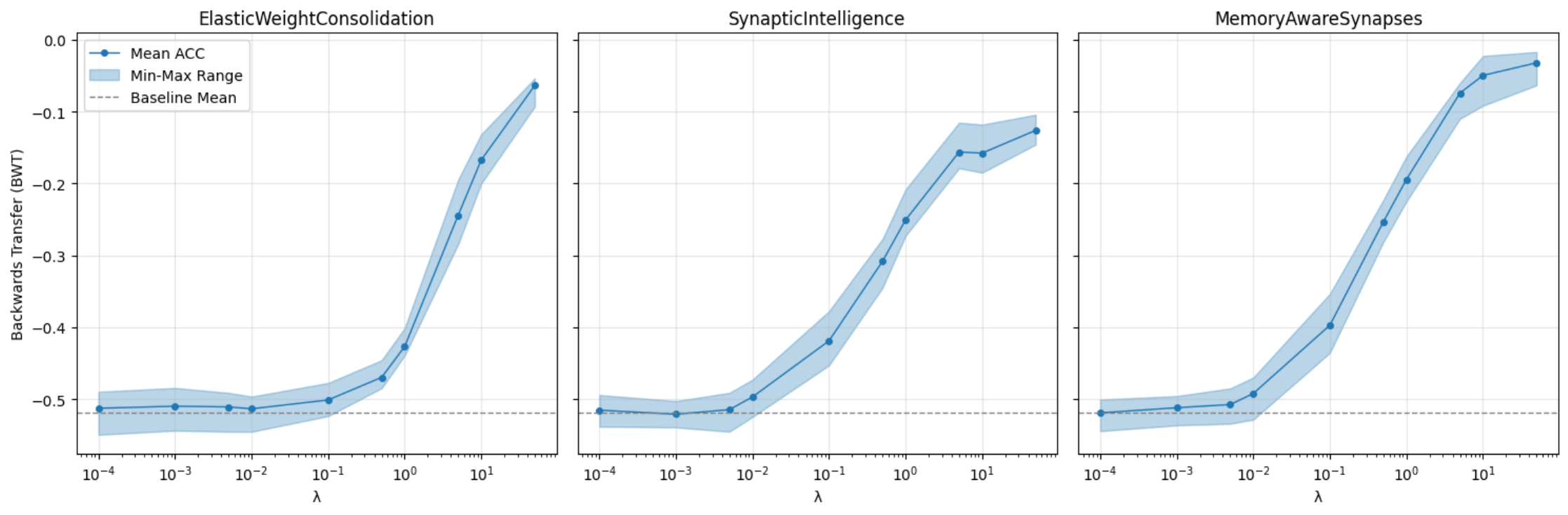}
            \caption{[DREAMER] Average BWT across seeds for each strategy with respect to increased regularisation ($\lambda$).}
            \label{fig:dreamer_bwt}
        \end{figure}
        \begin{figure}[hbt!]
            \centering
            \captionsetup{width=0.85\linewidth}
            \includegraphics[width=0.85\linewidth]{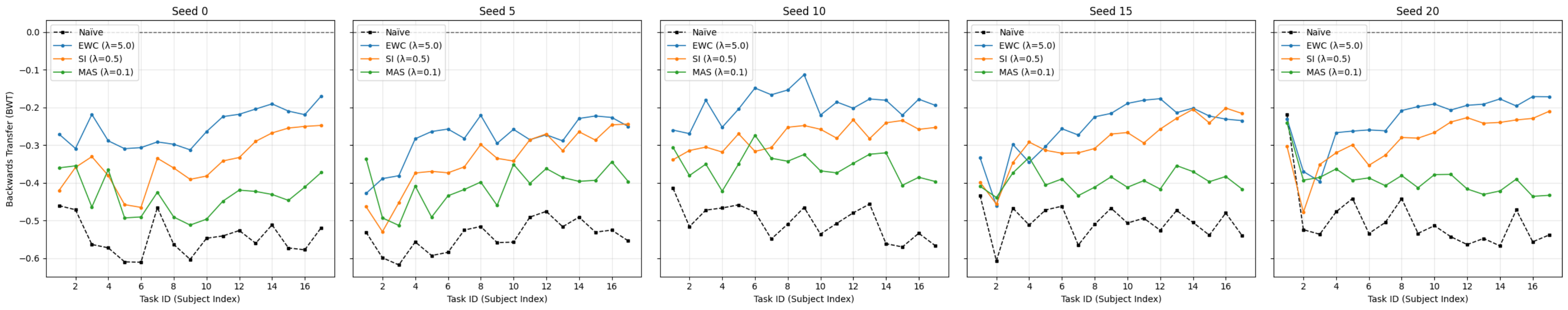}
            \caption{[DREAMER] BWT across tasks for each strategy and each seed, where EWC $\lambda = 5.0$, SI $\lambda = 0.5$, and MAS $\lambda = 0.1$.}
            \label{fig:dreamer_bwt_lambda1}
        \end{figure}

\item \textbf{No regularisation strength or method is able to reliably outperform the baseline on FWT.} FWT measures zero-shot accuracy relative to accuracy at random initialisation. On both datasets, FWT on each strategy sometimes outperforms and sometimes underperforms the baseline, suggesting that regularisation does not improve FWT. FWT patterns vary considerably across random seeds, suggesting subject-order highly influences knowledge transfer. A one sample t-test supports this observation, where the null hypothesis is that FWT is less than or equal to the baseline. For DREAMER: Elastic Weight Consolidation: $p=0.686$, Synaptic Intelligence: $p=0.664$, Memory Aware Synapses: $p=0.433$; and SEED: Elastic Weight Consolidation: $p=0.175$, Synaptic Intelligence: $p=0.546$, Memory Aware Synapses: $p=0.456$; all $p>0.05$ such that none reject the null hypothesis. It is clear that these methods do not reliably facilitate forward knowledge transfer.
        
    \begin{figure}[hbt!]
        \centering
        \captionsetup{width=0.85\linewidth}
        \includegraphics[width=0.85\linewidth]{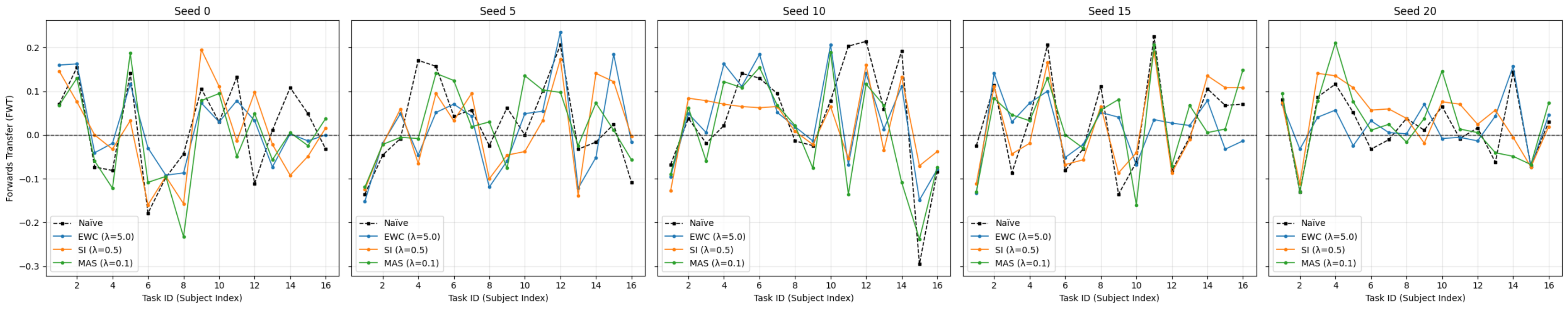}
        \caption{[DREAMER] FWT across seeds and strategies, where EWC $\lambda = 5.0$, SI $\lambda = 0.5$, and MAS $\lambda = 0.1$.}
        \label{fig:dreamer_fwt}
    \end{figure}
    \begin{figure}[hbt!]
        \centering
        \captionsetup{width=0.85\linewidth}
        \includegraphics[width=0.85\linewidth]{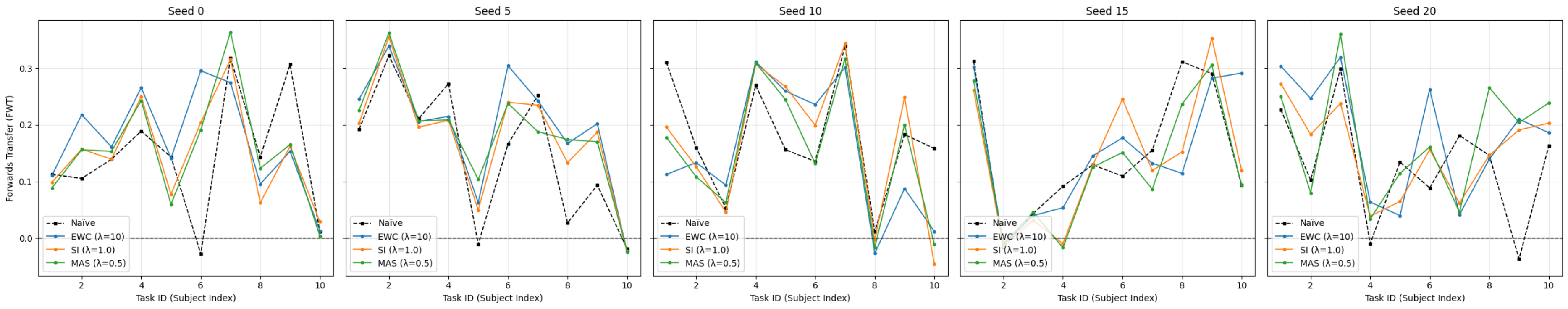}
        \caption{[SEED] FWT across seeds and strategies, where EWC $\lambda = 10.0$, SI $\lambda = 1.0$, and MAS $\lambda = 0.5$.}
        \label{fig:seed_fwt}
    \end{figure}


\item Table~\ref{tab:lambdas} reveals that no strategy is able to consistently outperform the baseline on unseen subject F1 score. It is evident that understanding the underlying mechanistic reasons that are influencing these observations is necessary.

\begin{table*}[hbt!]
\centering
\tiny
\caption{Performance at the given regularisation strengths per strategy and dataset, averaged across seeds.}
\label{tab:lambdas}
\begin{tabular}{llccccccc}
\toprule
\textbf{Dataset} & \textbf{Strategy} & $\boldsymbol{\lambda}$ & \textbf{Mean ACC} & \textbf{Final ACC} & \textbf{BWT} & \textbf{FWT} &\textbf{Unseen F1} \\
\midrule

\multirow{4}{*}{\textbf{DREAMER}} 
& Naive     & -    & 0.3877 ± 0.0104            & 0.2973 ± 0.0135           & -0.5191 ± 0.0230          & \textbf{0.0257 ± 0.0111}  & 0.2232 ± 0.0217 \\
& EWC       & 5.0  & \textbf{0.4400 ± 0.0188}   & \textbf{0.3491 ± 0.0338}  & \textbf{-0.2444 ± 0.0333} & 0.0219 ± 0.0116           & \textbf{0.2455 ± 0.0263} \\
& SI        & 0.5  & 0.4068 ± 0.0135            & 0.3374 ± 0.0199           & -0.3084 ± 0.0324          & 0.0215 ± 0.0142           & 0.2311 ± 0.0155 \\
& MAS       & 0.1  & 0.4209 ± 0.0094            & 0.3316 ± 0.0272           & -0.3973 ± 0.0300          & 0.0176 ± 0.0160           & 0.2295 ± 0.0330 \\

\midrule

\multirow{4}{*}{\textbf{SEED}}
& Naive     & -    & 0.5760 ± 0.0065            & \textbf{0.5240 ± 0.0253}  & -0.2994 ± 0.0194          & 0.1508 ± 0.0175   & \textbf{0.4577 ± 0.0502}\\
& EWC       & 10.0 & 0.5810 ± 0.0148            & 0.5215 ± 0.0245           & \textbf{-0.1426 ± 0.0146} & \textbf{0.1710 ± 0.0189}          & 0.4423 ± 0.0599\\
& SI        & 1.0  & 0.5861 ± 0.0168            & 0.5093 ± 0.0375           & -0.1929 ± 0.0255          & 0.1582 ± 0.0156           & 0.4262 ± 0.0527\\
& MAS       & 0.5  & \textbf{0.5941 ± 0.0139}   & 0.5014 ± 0.0464           & -0.2090 ± 0.0204          & 0.1595 ± 0.0218           & 0.4218 ± 0.0376\\

\bottomrule
\end{tabular}
\end{table*}

\end{itemize}

\subsection{H1a: \emph{(EWC)} Empirical Fisher Information deviates from the Fisher Information under smaller sample sizes; and the Fisher Information deviates from the expected Hessian under model misspecification.}
\begin{figure}[hbt!]
    \centering
    \includegraphics[width=0.9\linewidth]{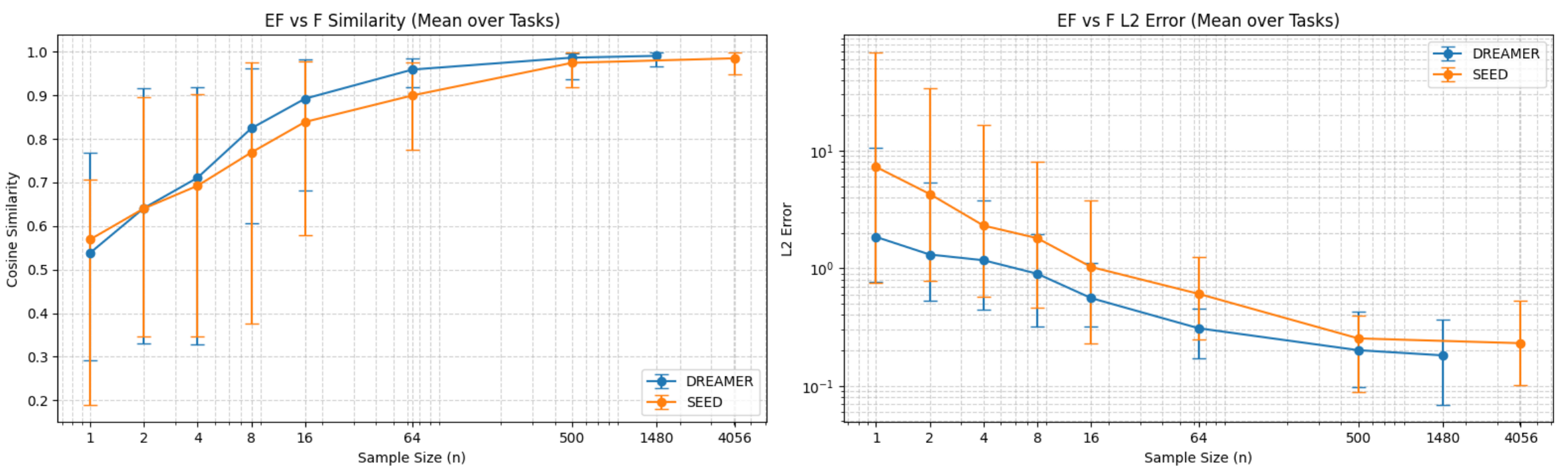}
    \caption{Effect of increasing the sample size on convergence of empirical Fisher Information to the true Fisher Information.}
    \label{fig:fisher_convergence}
\end{figure}

Section~\ref{sec:fisher} identified that the dataset must be sufficiently large for the empirical Fisher Information to approximate the Fisher Information. Figure~\ref{fig:fisher_convergence} shows that as the sample size increases, the empirical Fisher Information converges to the Fisher Information in both cosine similarity and relative L2-error. This suggests that at least approximately 500 samples in our data must be used by the empirical Fisher Information to approximate the true Fisher Information.
While the empirical Fisher Information approximates the true Fisher Information under a large sample size, the true Fisher Information can be a poor proxy of the expected Hessian as a result of poor convergence caused by noise and model misspecification.

\subsection{H1b: \emph{(SI)} The path integral inflates under stochastic noise and EEG signal noise (incl.\ intra-subject variability).}
\begin{figure}[hbt!]
    \centering
    \includegraphics[width=0.9\linewidth]{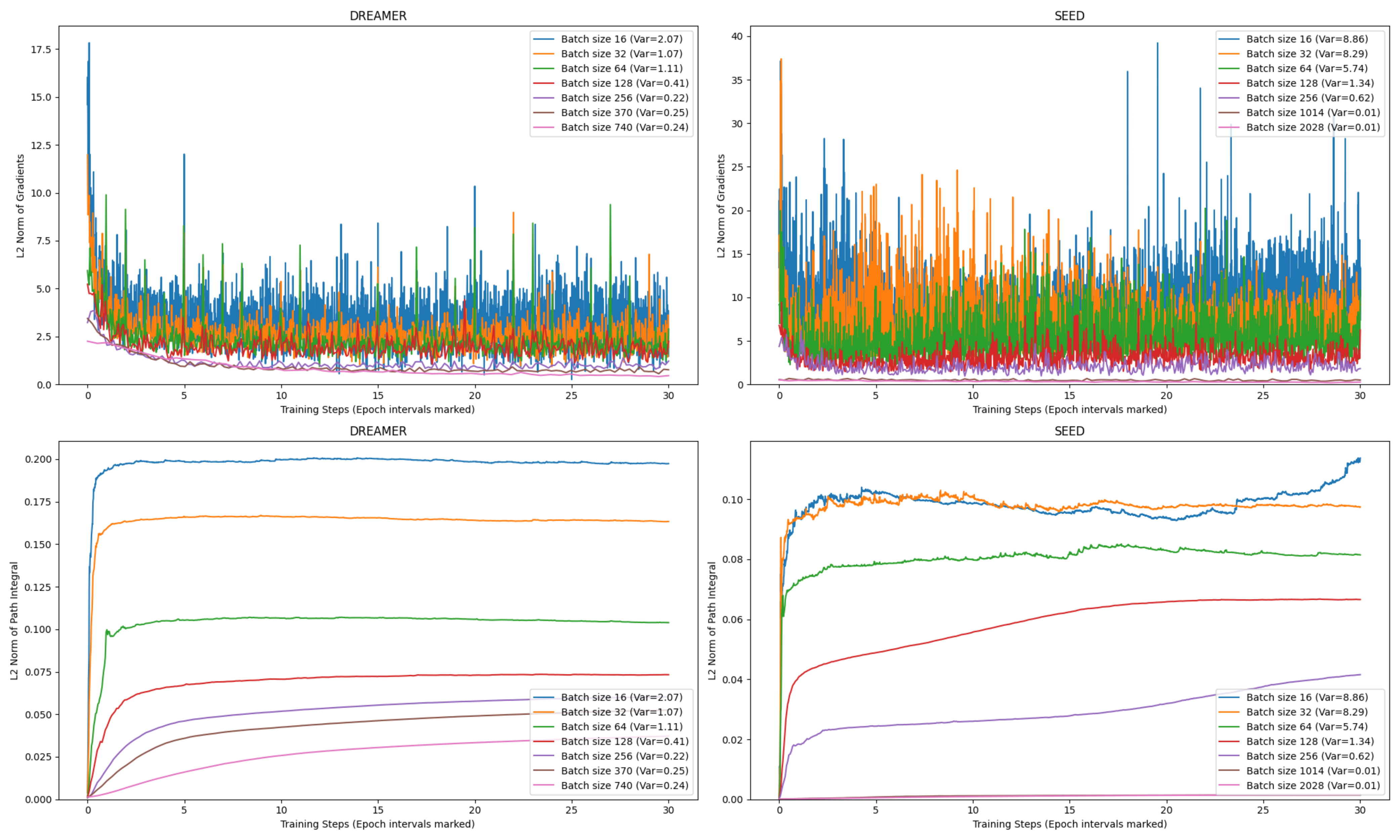}
    \caption{Effect of decreasing the batch size on inflation of the path integral in synaptic intelligence.}
    \label{fig:h1_si_path_integral}
\end{figure}

Figure~\ref{fig:h1_si_path_integral} shows gradient variance increases under smaller batch sizes, which corresponds to an increase in magnitude of the path integral. While much of this inflation is caused by sampling variance from smaller batches, the effect demonstrates the sensitivity of SI's path integral to gradient noise, a property typically worse with low-SNR EEG signals. A Pearson correlation test indicates both strong and statistically significant correlation between gradient variance and the path integral for both DREAMER ($\rho = 0.939$, $p = 0.002$) and SEED ($\rho = 0.909$, $p = 0.005$). 

\subsection{H1c: \emph{(MAS)} Importance estimates are more robust under stochastic noise.}
\begin{figure}[hbt!]
    \centering
    \includegraphics[width=0.9\linewidth]{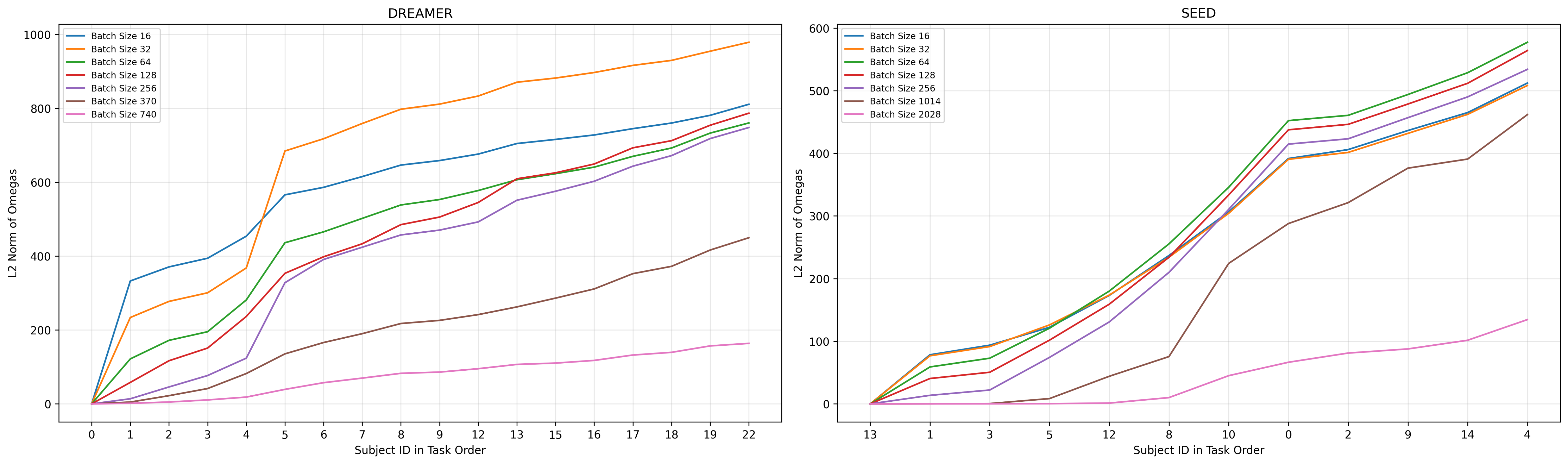}
    \caption{Effect of decreasing the batch size on magnitude of importance estimates in memory aware synapses.}
    \label{fig:mas_intra_sample_size}
\end{figure}

According to Figure~\ref{fig:mas_intra_sample_size}, when importance is accumulated using a smaller batch size during training, there is a moderate correlation between batch size and Omega on DREAMER ($\rho = 0.656$, $p = 0.000$) and no correlation on SEED ($\rho = -0.093$, $p = 0.399$). This adheres to expected behaviour, as MAS is more robust to stochastic noise.

\subsection{H2: Subsequent subjects require conflicting parameter updates, leading to gradient interference rather than convergence, limiting generalisability.}
\begin{figure}[hbt!]
    \centering
    \includegraphics[width=0.9\linewidth]{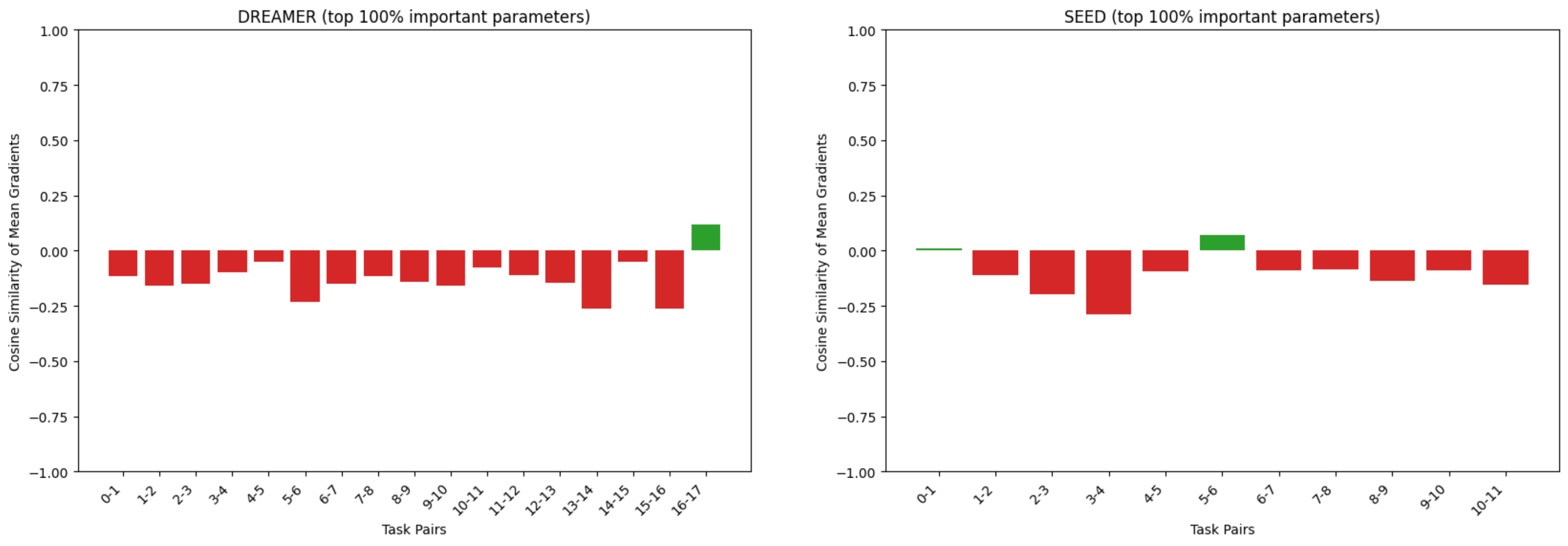}
    \captionsetup{width=0.9\linewidth}
    \caption{[Naïve] Cosine similarity of all gradients between subsequent tasks without regularisation. The $y$-axis is fixed to the full cosine similarity range $[-1, 1]$ to preserve interpretability and comparability across task pairs and methods. No per-bar parameter-count labels are shown, as all parameters are shared between tasks (100\% overlap) for the naïve strategy.}
    \label{fig:cosine_naive}
\end{figure}

\begin{figure}[hbt!]
    \centering
    \includegraphics[width=0.9\linewidth]{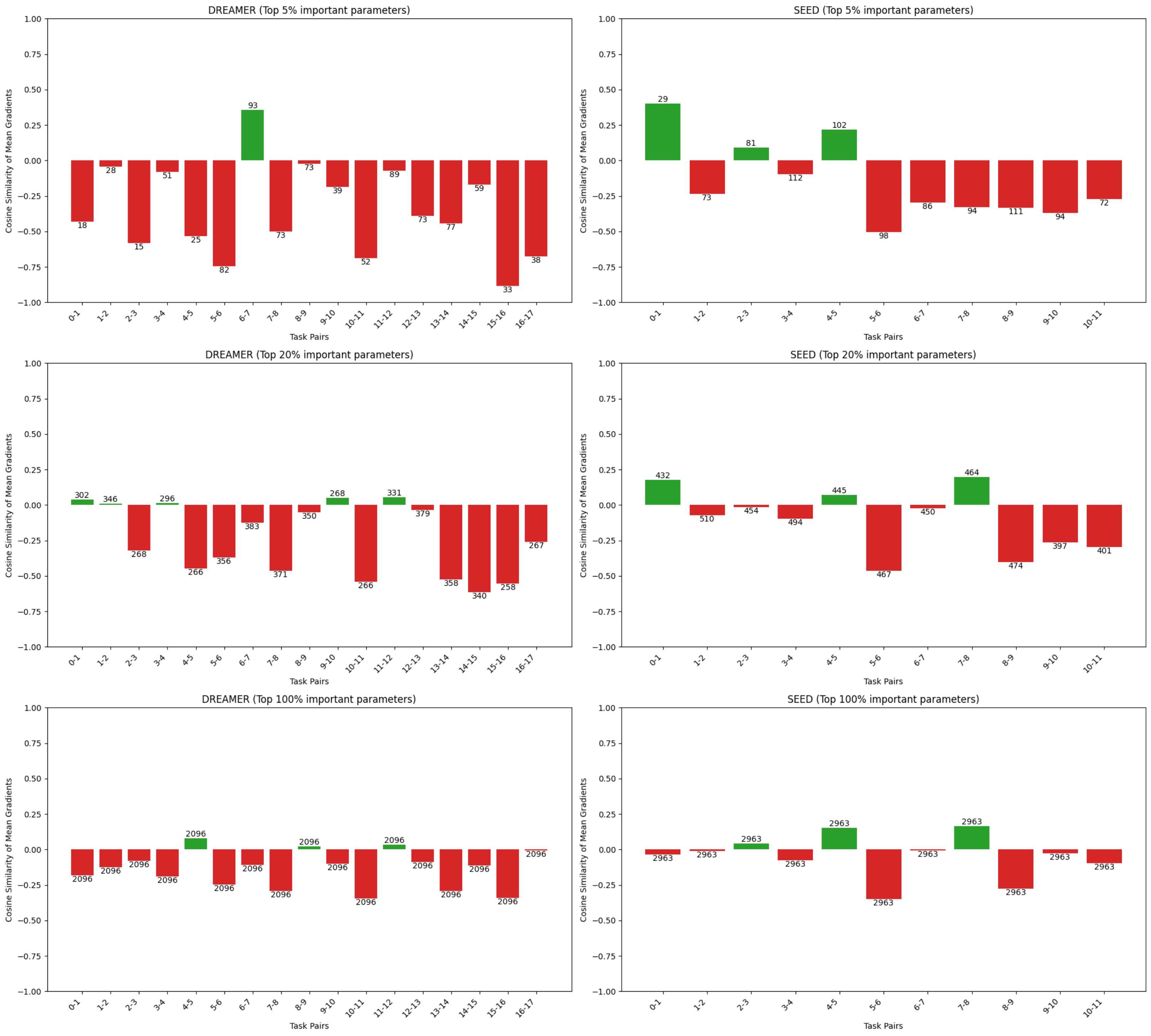}
    \captionsetup{width=0.9\linewidth}
    \caption{[EWC] Cosine similarity of gradients of parameters in top 5\%, 20\%, 100\% of both tasks in task pair. The label above or below each bar indicates the number of parameters that were in this overlap.}
    \label{fig:ewc_cosine}
\end{figure}

Catastrophic forgetting can be measured by gradient interference between tasks, as is illustrated by Figure~\ref{fig:cosine_naive}. When subsequent tasks are substantially dissimilar, gradients at parameters which are important to both tasks might interfere, such that constraining plasticity on these parameters actively degrades performance of the subsequent task. It is evident that gradient interference persists despite regularisation (Figure~\ref{fig:ewc_cosine}) constraining gradient magnitudes. Not only can this limit performance if the model is constrained to a poor compromise, but this will also limit generalisability, as gradients between tasks do not converge to a shared representation.

\subsection{H3: Importance accumulates across tasks, constraining plasticity in the network early, limiting scalability.}
\label{sec:H3}
By the definition of importance ($\Omega$) for each method, importance accumulates across tasks, as subsequent tasks are learnt. Omega values generally grow linearly, with less constant growths observed in EWC due to the exponential moving average nature of its importance estimates (Figure~\ref{fig:si_h2_omegacurve} and \ref{fig:ewc_h2_omegacurve}, \ref{fig:mas_h2_omegacurve} in the Appendix).
This persistent increase in Omega is unbounded, which will eventually lead to over-constraining of the network. While the rate of accumulation can decrease as regularisation constraints affect the gradients, the monotonic growth in importance estimates remains unscalable for long task sequences.
\begin{figure}[hbt!]
    \centering
    \captionsetup{width=0.9\linewidth}
    \includegraphics[width=0.9\linewidth]{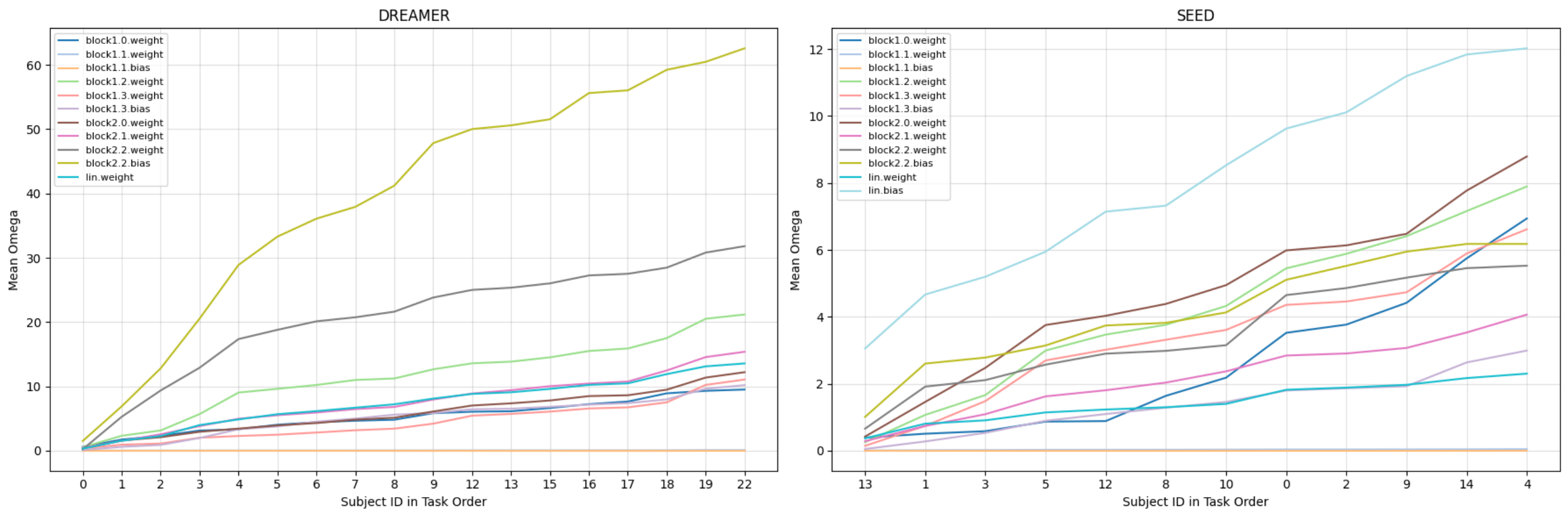}
    \caption{[SI] Mean importance per named parameter across tasks. Importance is measured by the SI heuristic: the cumulative path integral.}
    \label{fig:si_h2_omegacurve}
\end{figure}

This importance accumulation is reflected by a decrease in network plasticity as parameter changes become smaller across subsequent tasks (Figure~\ref{fig:si_h2_meandelta} and \ref{fig:ewc_h2_meandelta}, \ref{fig:mas_h2_meandelta} in the Appendix). Each approach shows a sharp decrease in parameter change, with only minor changes to parameters towards the end of learning.
\begin{figure}[hbt!]
    \centering
    \captionsetup{width=1\linewidth}
    \includegraphics[width=0.9\linewidth]{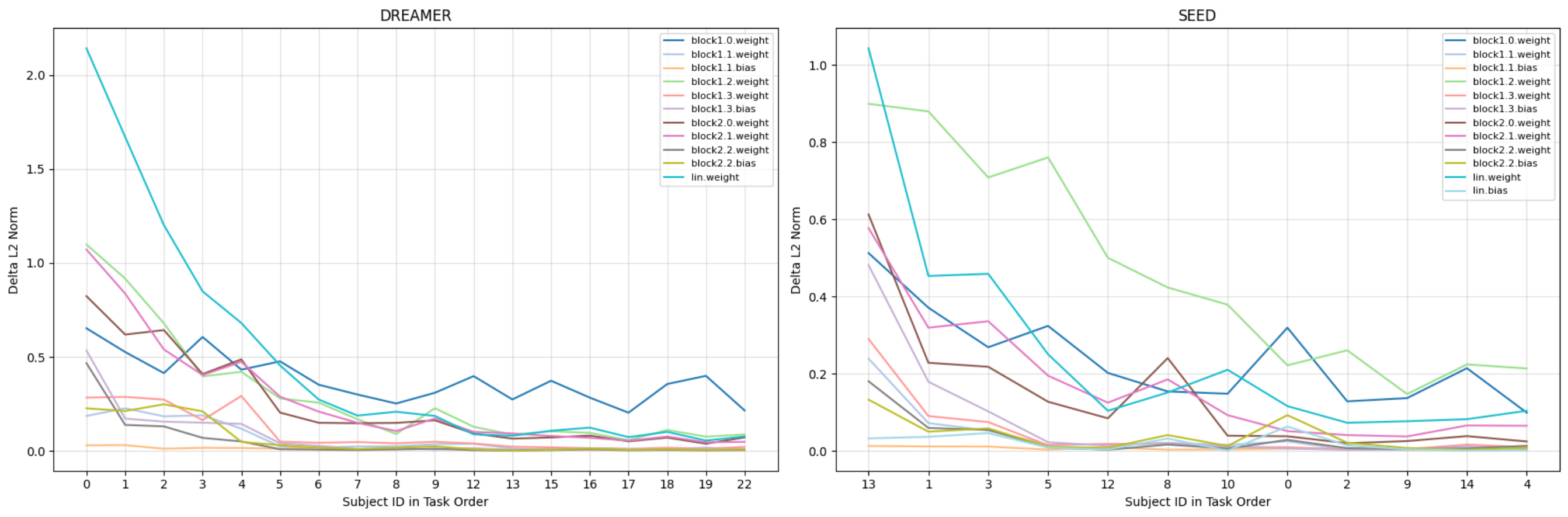}
    \caption{[SI] Magnitude of parameter changes (L2 norm) across tasks, per named parameter.}
    \label{fig:si_h2_meandelta}
\end{figure}

Figure~\ref{fig:si_h2_deltascatter} (and \ref{fig:ewc_h2_deltascatter}, \ref{fig:mas_h2_deltascatter} in the Appendix) illustrate this correlation between increased importance and decreased parameter changes, as is expected behaviour for regularisation-based approaches. This correlation is statistically significant for EWC, SI, and MAS, across DREAMER and SEED, with the exception of EWC on SEED. The Pearson-correlation coefficients and their $p$-value for this are found in the title of each plot.
\begin{figure}[hbt!]
    \centering
    \captionsetup{width=1\linewidth}
    \includegraphics[width=0.9\linewidth]{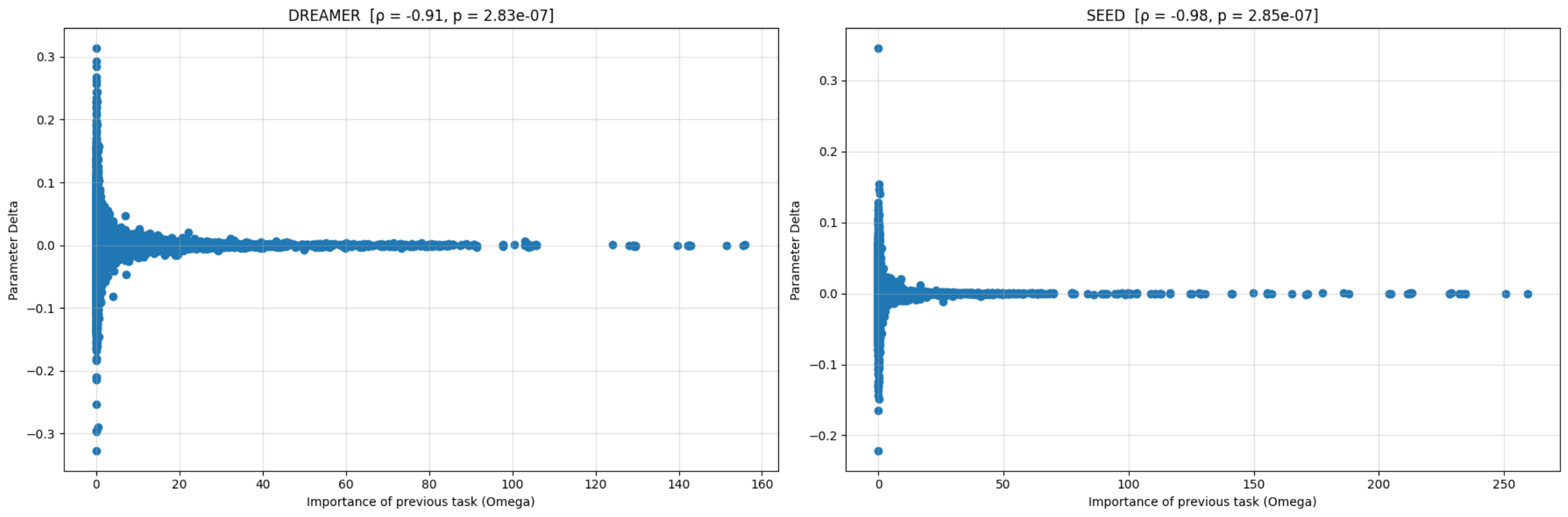}
    \caption{[SI] Correlation between importance of the previous task and the parameter change in the subsequent task.}
    \label{fig:si_h2_deltascatter}
\end{figure}

This demonstrates that as the subject sequence becomes longer, parameter changes become increasingly smaller, until important parameters are frozen to such an extent that they are no longer able to meaningfully adapt to incoming subjects.

\subsection{H4: Subject order impacts which parameters are constrained early, making each method highly sensitive to random seed.}
When subject order is changed, earlier subjects are less constrained by weight-regularisation. This means they influence the optimal parameters for the system considerably more. As a result, different subject orders cause variable performance to unseen subjects (Tables \ref{tab:dreamer_h3}, \ref{tab:seed_h3}). Regularisation-based approaches can therefore not guarantee consistent performance across different subject orders, as determined by random seeds.

\begin{table}[hbt!]
\small
\centering
\begin{tabular}{c|ccccc|c}
             & \textbf{Shuffle 1} & \textbf{Shuffle 2} & \textbf{Shuffle 3} & \textbf{Shuffle 4} & \textbf{Shuffle 5} & \textbf{Mean} \\
             \toprule
\textbf{Naïve}  & 0.2280      & 0.1683    & 0.2001    & 0.3036    & 0.1911  & 0.2182    \\
\textbf{EWC}    & 0.2396      & 0.1568    & 0.1707    & 0.2752    & 0.2422  & 0.2169    \\
\textbf{SI}     & 0.2571      & 0.1410    & 0.2014    & 0.1789    & 0.1577  & 0.1872      \\
\textbf{MAS}    & 0.1904      & 0.1119    & 0.2465    & 0.1566    & 0.2908  & 0.1993  
\end{tabular}
\captionsetup{width=0.7\linewidth}
\caption{[DREAMER] F1-score for unseen subject 11 when random seed, train and unseen subjects are kept the same, but train subject order is shuffled. Random accuracy is 0.25 for 4-class classification.}
\label{tab:dreamer_h3}
\end{table}
\begin{table}[hbt!]
\small
\centering
\begin{tabular}{c|ccccc|c}
             & \textbf{Shuffle 1} & \textbf{Shuffle 2} & \textbf{Shuffle 3} & \textbf{Shuffle 4} & \textbf{Shuffle 5} & \textbf{Mean} \\
             \toprule
\textbf{Naïve}  & 0.4555        & 0.3958    & 0.3954    & 0.5862    & 0.5819    & 0.4829   \\
\textbf{EWC}    & 0.5903        & 0.5054    & 0.5148    & 0.4788    & 0.3241    & 0.4827    \\
\textbf{SI}     & 0.4338        & 0.4542    & 0.5346    & 0.5148    & 0.5400    & 0.4954    \\
\textbf{MAS}    & 0.4124        & 0.5215    & 0.3342    & 0.4744    & 0.4654    & 0.4416   
\end{tabular}
\captionsetup{width=0.7\linewidth}
\caption{[SEED] F1-score for unseen subject 10 when random seed, train and unseen subjects are kept the same, but train subject order is shuffled. Random accuracy is 0.33 for 3-class classification.}
\label{tab:seed_h3}
\end{table}

Tables~\ref{tab:dreamer_h3} and \ref{tab:seed_h3} show both high variation in performance across a strategy, as well as the inability for any strategy to consistently outperform the naïve baseline. This is evident by some strategies outperforming the baseline on one shuffle but not on another. A visual interpretation of these results across additional unseen subjects is provided in Appendix~\ref{sec:appendix-h4}, under Figures~~\ref{fig:dreamer_shuffle} and \ref{fig:seed_shuffle}. This instability limits the reliability of regularisation-based continual learning for EEG-based emotion classification, where subject order is uncertain and inter-subject variability is unavoidable.

\section{Discussion}
\label{sec:discussion}

Regularisation-based continual learning approaches are designed to mitigate catastrophic forgetting. They prioritise remembering (backward transfer) over performing well on new tasks (forward transfer), indicating a \textbf{conceptual misalignment of objectives} between these methods and our primary objective. The core issue is that remembering is only useful to our objective if past knowledge supports future performance. If subsequent tasks require conflicting parameter updates, the retained information does not contribute to performance of the next task. The negative cosine similarity between gradients of subsequent subjects is a clear indicator of this underlying gap, showing the extent to which the model's constrained parameters are incompatible with the new task. While the stability-plasticity trade-off can be adjusted via regularisation-strength, this becomes less relevant as no amount of tuning can overcome this fundamental gap. While regularisation-based methods may successfully improve performance on past tasks as intended, they are unable to consistently generalise to unseen tasks.

\textbf{Importance estimates become less reliable under stochastic noise and EEG signal noise}, as they can adapt to noise more than true curvature of the loss landscape. For EWC, while the empirical Fisher Information
is the most commonly used approximator, it is inherently flawed and known to capture gradient noise, which can be amplified by intra-task variability. This may limit stable convergence at task boundaries, rendering importance estimates less reliable. Similarly, the path integral in SI is skewed by both stochastic batch noise and sample-driven gradient fluctuations, leading to overestimated importance. While MAS is more robust to stochastic noise by estimating importance based on gradients of the model outputs rather than gradients of the loss during training, it remains an imperfect heuristic under high noise. For each method, the importance estimates under high intra-subject variability and EEG signal noise are less reliable, which can cause the wrong parameters to be excessively regularised.

Regularisation-based CL approaches are not designed for a large or theoretically infinite stream of tasks, making them inherently \textbf{unscalable}. EWC's penalty was originally derived for two-task scenarios \cite{huszar_note_2018, kirkpatrick_overcoming_2017}, with SI and MAS also designed for few-task settings. As the number of tasks increases, the penalty accumulates linearly. This is both identified by Schwarz et al.\  \cite{schwarz_progress_2018}, as well as our empirical findings (Section~\ref{sec:H3}). While Online EWC attempts to mitigate this by introducing a decay factor to balance past and present task importance, the core problem of penalty accumulation persists. As the number of tasks grows, the quadratic penalties of EWC, SI, and MAS try to stay within the optimal parameter regions of an increasing number of tasks, over-constraining the network and resulting in early loss of plasticity and impaired learning of new tasks. The network parameters are essentially frozen, making learning from new subjects/tasks negligible. This early freezing creates a strong sensitivity to subject-order and therefore random seed with later subjects having negligible influence. While this stability might be acceptable if the model has reached a global optimum, such a static system defeats the initial premise of continual learning, which is based on a constant need to adapt. 

While \textbf{lightweight fixes} such as Omega decay, Omega clipping, or use of more more robust heuristics may prevent early freezing of the network and alleviate some of the identified limitations, the fundamental stability-plasticity trade-off where the network is either too constrained or cannot retain knowledge, persists --- along with the inherent inability to transfer knowledge forward.

\section{Limitations}
\label{sec:limitations}
Despite its contributions, this work has several limitations.

Firstly, EEG signal preprocessing in this study is limited by a lack of expert knowledge on manual artefact removal. This study relied on an automatic artefact removal process using a z-scored kurtosis threshold within a FastICA transform, rather than manually going through the data using tools such as ICLabel in EEGLab or the MNE Python library. The kurtosis is a rather outdated technique, although still accessible on EEGLab. Consequently, some artefacts may still be present, aggravating the already low signal-to-noise ratio.

Further, the within-trial train-test splitting approach following TorchEEG's    \verb+train_test_split_groupby_trial+ introduces minimal data leakage as adjacent segments at just the split boundary are temporally correlated. This was a necessary trade-off as the continual learning setting does not allow for a typical k-fold approach such that test-performance was insufficient otherwise for computing continual learning metrics such as BWT and FWT. These test metrics serve only to measure knowledge retention and transfer rather than to support generalisability claims.

\section{Future Work}
This study has thoroughly evaluated the limitations of regularisation-based approaches for EEG-based emotion classification. It is therefore sensible to direct future research into more promising directions with more robust and adaptive approaches.



An exciting area for exploration is \textbf{meta-learning}, which enables a model to learn-to-learn \cite{wang_comprehensive_2024}, without relying on replay. Techniques like Model-Agnostic Meta-Learning (MAML) \cite{finn_model-agnostic_2017} are particularly relevant, as they are designed for rapid adaptation to new tasks with minimal data \cite{javaid_model_2023}, an obstacle identified for regularisation-based approaches which need a large sample size for accurate estimates. More recent advancements, such as neuromimetic metaplasticity \cite{cho_neuromimetic_2025}, adaptive meta-optimizers \cite{duan_online_2024}, and metaplastic synapses for EEG signals \cite{aguilar_continuous_2025} further emphasise the potential of meta-learning in enabling fast adaptation across diverse tasks and subjects.

Lastly, the development of \textbf{EEG Foundation Models (EEG-FMs)} is a remarkable step forward in leveraging EEG data \cite{lai_simple_2025}. By pre-training on a large number of diverse datasets, these models learn underlying neural patterns, facilitating generalisability across subjects, BCI paradigms, and a range of downstream tasks. Consequently, leveraging a pre-trained EEG-FM as a starting point for fine-tuning could potentially mitigate the instability and forgetting observed when training on one subject at a time. Foundation models are especially promising for continual learning beyond domain-incremental learning and into task-incremental learning, where one model supports multiple BCI paradigms.

\section{Conclusion}
This study identified a fundamental misalignment of objectives between regularisation-based approaches (retaining past knowledge) and EEG-based emotion classification (improving performance on future subjects). Through both theoretical and empirical investigation, we demonstrated that these approaches are inherently limited by design, further compounded by the effects of high inter- and intra-subject variability. This was demonstrated by the evaluation of four hypotheses, which concluded that (1) importance estimates are highly sensitive to stochastic noise and intra-task variability, (2) subsequent subjects required conflicting parameter updates which impairs knowledge transfer, (3) importance accumulates across tasks, overly restricting parameter plasticity, which (4) makes forward performance and generalisation on unseen subjects less reliable. This study therefore finds that despite their popularity as baseline methods in EEG-based continual learning, regularisation-based approaches provide limited utility to subject-independent EEG-based emotion classification. Novel techniques such as meta-learning and foundation models provide exciting directions for future research to tackle this paradigm.

\clearpage
\bibliographystyle{ACM-Reference-Format}
\bibliography{strings,refs}

@article{zheng2015investigating, title={Investigating Critical Frequency Bands and Channels for {EEG}-based Emotion Recognition with Deep Neural Networks}, author={Zheng, Wei-Long and Lu, Bao-Liang}, journal={IEEE Transactions on Autonomous Mental Development}, doi={10.1109/TAMD.2015.2431497}, year={2015}, volume={7}, number={3}, pages={162-175}, publisher={IEEE} }

@inproceedings{duan2013differential, title={Differential entropy feature for {EEG}-based emotion classification}, author={Duan, Ruo-Nan and Zhu, Jia-Yi and Lu, Bao-Liang}, booktitle={6th International IEEE/EMBS Conference on Neural Engineering (NER)}, pages={81--84}, year={2013}, organization={IEEE} }

@ARTICLE{dreamer,
  author={Katsigiannis, Stamos and Ramzan, Naeem},
  journal={IEEE Journal of Biomedical and Health Informatics}, 
  title={DREAMER: A Database for Emotion Recognition Through EEG and ECG Signals From Wireless Low-cost Off-the-Shelf Devices}, 
  year={2018},
  volume={22},
  number={1},
  pages={98-107},
  doi={10.1109/JBHI.2017.2688239}}

@article{huszar_note_2018,
    author = {Ferenc Huszár},
    title = {Note on the quadratic penalties in elastic weight consolidation},
    journal = {Proceedings of the National Academy of Sciences},
    volume = {115},
    number = {11},
    pages = {E2496-E2497},
    year = {2018},
    doi = {10.1073/pnas.1717042115},
    URL = {https://www.pnas.org/doi/abs/10.1073/pnas.1717042115},
    eprint = {https://www.pnas.org/doi/pdf/10.1073/pnas.1717042115}
}

@article{Kamble2023,
  author    = {Kamble, Kranti and Sengupta, Joydeep},
  title     = {A comprehensive survey on emotion recognition based on electroencephalograph (EEG) signals},
  journal   = {Multimedia Tools and Applications},
  volume    = {82},
  number    = {18},
  pages     = {27269--27304},
  year      = {2023},
  month     = {July},
  doi       = {10.1007/s11042-023-14489-9},
  url       = {https://doi.org/10.1007/s11042-023-14489-9},
  issn      = {1573-7721}
}

@article{DELORME20049,
title = {EEGLAB: an open source toolbox for analysis of single-trial EEG dynamics including independent component analysis},
journal = {Journal of Neuroscience Methods},
volume = {134},
number = {1},
pages = {9-21},
year = {2004},
issn = {0165-0270},
doi = {https://doi.org/10.1016/j.jneumeth.2003.10.009},
url = {https://www.sciencedirect.com/science/article/pii/S0165027003003479},
author = {Arnaud Delorme and Scott Makeig}
}

@misc{kingma_adam_2017,
	title = {Adam: {A} {Method} for {Stochastic} {Optimization}},
	shorttitle = {Adam},
	url = {http://arxiv.org/abs/1412.6980},
	doi = {10.48550/arXiv.1412.6980},
	urldate = {2025-07-28},
	publisher = {arXiv},
	author = {Kingma, Diederik P. and Ba, Jimmy},
	month = jan,
	year = {2017},
	note = {arXiv:1412.6980 [cs]},
}

@INPROCEEDINGS{7391935,
  author={Kazi Aminul Islam and Tcheslavski, Gleb V.},
  booktitle={2015 2nd International Conference on Electrical Information and Communication Technologies (EICT)}, 
  title={Independent Component Analysis for EOG artifacts minimization of EEG signals using kurtosis as a threshold}, 
  year={2015},
  volume={},
  number={},
  pages={137-142},
  doi={10.1109/EICT.2015.7391935}}

@article{lawhern_eegnet_2018,
doi = {10.1088/1741-2552/aace8c},
url = {https://doi.org/10.1088/1741-2552/aace8c},
year = {2018},
month = {jul},
publisher = {IOP Publishing},
volume = {15},
number = {5},
pages = {056013},
author = {Lawhern, Vernon J and Solon, Amelia J and Waytowich, Nicholas R and Gordon, Stephen M and Hung, Chou P and Lance, Brent J},
title = {EEGNet: a compact convolutional neural network for EEG-based brain–computer interfaces},
journal = {Journal of Neural Engineering}
}

@article{tao_eeg-based_2023,
author = {Tao, Wei and Li, Chang and Song, Rencheng and Cheng, Juan and Liu, Yu and Wan, Feng and Chen, Xun},
title = {EEG-Based Emotion Recognition via Channel-Wise Attention and Self Attention},
year = {2023},
issue_date = {Jan.-March 2023},
publisher = {IEEE Computer Society Press},
address = {Washington, DC, USA},
volume = {14},
number = {1},
issn = {1949-3045},
url = {https://doi.org/10.1109/TAFFC.2020.3025777},
doi = {10.1109/TAFFC.2020.3025777},
journal = {IEEE Trans. Affect. Comput.},
month = jan,
pages = {382–393},
numpages = {12}
}

@Article{hu_eeg-based_2022,
AUTHOR = {Hu, Zhangfang and Chen, Libujie and Luo, Yuan and Zhou, Jingfan},
TITLE = {EEG-Based Emotion Recognition Using Convolutional Recurrent Neural Network with Multi-Head Self-Attention},
JOURNAL = {Applied Sciences},
VOLUME = {12},
YEAR = {2022},
NUMBER = {21},
ARTICLE-NUMBER = {11255},
URL = {https://www.mdpi.com/2076-3417/12/21/11255},
ISSN = {2076-3417},
DOI = {10.3390/app122111255}
}

@article{wang_comprehensive_2024,
author = {Wang, Liyuan and Zhang, Xingxing and Su, Hang and Zhu, Jun},
title = {A Comprehensive Survey of Continual Learning: Theory, Method and Application},
year = {2024},
issue_date = {Aug. 2024},
publisher = {IEEE Computer Society},
address = {USA},
volume = {46},
number = {8},
issn = {0162-8828},
url = {https://doi.org/10.1109/TPAMI.2024.3367329},
doi = {10.1109/TPAMI.2024.3367329},
journal = {IEEE Trans. Pattern Anal. Mach. Intell.},
month = aug,
pages = {5362–5383},
numpages = {22}
}

@article{van_de_ven_three_2022,
	title = {Three types of incremental learning},
	volume = {4},
	copyright = {2022 The Author(s)},
	issn = {2522-5839},
	url = {https://www.nature.com/articles/s42256-022-00568-3},
	doi = {10.1038/s42256-022-00568-3},
	language = {en},
	number = {12},
	urldate = {2025-05-21},
	journal = {Nature Machine Intelligence},
	author = {van de Ven, Gido M. and Tuytelaars, Tinne and Tolias, Andreas S.},
	month = dec,
	year = {2022},
	note = {Publisher: Nature Publishing Group},
	pages = {1185--1197},
}

@article{kirkpatrick_overcoming_2017,
author = {James Kirkpatrick  and Razvan Pascanu  and Neil Rabinowitz  and Joel Veness  and Guillaume Desjardins  and Andrei A. Rusu  and Kieran Milan  and John Quan  and Tiago Ramalho  and Agnieszka Grabska-Barwinska  and Demis Hassabis  and Claudia Clopath  and Dharshan Kumaran  and Raia Hadsell },
title = {Overcoming catastrophic forgetting in neural networks},
journal = {Proceedings of the National Academy of Sciences},
volume = {114},
number = {13},
pages = {3521-3526},
year = {2017},
doi = {10.1073/pnas.1611835114},
URL = {https://www.pnas.org/doi/abs/10.1073/pnas.1611835114},
eprint = {https://www.pnas.org/doi/pdf/10.1073/pnas.1611835114}}

@InProceedings{zenke_continual_2017,
  title = 	 {Continual Learning Through Synaptic Intelligence},
  author =       {Friedemann Zenke and Ben Poole and Surya Ganguli},
  booktitle = 	 {Proceedings of the 34th International Conference on Machine Learning},
  pages = 	 {3987--3995},
  year = 	 {2017},
  editor = 	 {Precup, Doina and Teh, Yee Whye},
  volume = 	 {70},
  series = 	 {Proceedings of Machine Learning Research},
  month = 	 {06--11 Aug},
  publisher =    {PMLR},
  url = 	 {https://proceedings.mlr.press/v70/zenke17a.html}
}

@Article{li_continual_2024,
AUTHOR = {Li, Ao and Li, Huayu and Yuan, Geng},
TITLE = {Continual Learning with Deep Neural Networks in Physiological Signal Data: A Survey},
JOURNAL = {Healthcare},
VOLUME = {12},
YEAR = {2024},
NUMBER = {2},
ARTICLE-NUMBER = {155},
URL = {https://www.mdpi.com/2227-9032/12/2/155},
PubMedID = {38255045},
ISSN = {2227-9032},
DOI = {10.3390/healthcare12020155}
}

@misc{lai_simple_2025,
	title = {A {Simple} {Review} of {EEG} {Foundation} {Models}: {Datasets}, {Advancements} and {Future} {Perspectives}},
	shorttitle = {A {Simple} {Review} of {EEG} {Foundation} {Models}},
	url = {http://arxiv.org/abs/2504.20069},
	doi = {10.48550/arXiv.2504.20069},
	urldate = {2025-08-27},
	publisher = {arXiv},
	author = {Lai, Junhong and Wei, Jiyu and Yao, Lin and Wang, Yueming},
	month = apr,
	year = {2025},
	note = {arXiv:2504.20069 [cs]},
}

@INPROCEEDINGS{javaid_model_2023,
  author={Javaid, Muhammad Hanzla and Shah, Irfan Ali and Javaid, Muhammad Sharjeel and Bin Irshad, Usama and Halim, Zahid},
  booktitle={2023 IEEE IAS Global Conference on Emerging Technologies (GlobConET)}, 
  title={Model Agnostic Meta Learning for EEG Classification: Multitask Approach}, 
  year={2023},
  volume={},
  number={},
  pages={1-4},
  doi={10.1109/GlobConET56651.2023.10150186}}

@InProceedings{finn_model-agnostic_2017,
  title = 	 {Model-Agnostic Meta-Learning for Fast Adaptation of Deep Networks},
  author =       {Chelsea Finn and Pieter Abbeel and Sergey Levine},
  booktitle = 	 {Proceedings of the 34th International Conference on Machine Learning},
  pages = 	 {1126--1135},
  year = 	 {2017},
  editor = 	 {Precup, Doina and Teh, Yee Whye},
  volume = 	 {70},
  series = 	 {Proceedings of Machine Learning Research},
  month = 	 {06--11 Aug},
  publisher =    {PMLR},
  url = 	 {https://proceedings.mlr.press/v70/finn17a.html}
}

@misc{ahmad_robust_2025,
	title = {Robust {Emotion} {Recognition} via {Bi}-{Level} {Self}-{Supervised} {Continual} {Learning}},
	url = {http://arxiv.org/abs/2505.10575},
	doi = {10.48550/arXiv.2505.10575},
	urldate = {2025-05-22},
	publisher = {arXiv},
	author = {Ahmad, Adnan and Nakisa, Bahareh and Rastgoo, Mohammad Naim},
	month = may,
	year = {2025},
}

@misc{EU_AI_act,
title = {Regulation (EU) 2024/1689 Artificial Intelligence Act },
author = {European Union},
year = 2024,
month = jun,
url = {http://data.europa.eu/eli/reg/2024/1689/oj},
note = {Regulation (EU) 2024/1689 of the European Parliament and of the Council of 13 June 2024 laying down harmonised rules on artificial intelligence and amending Regulations (EC) No 300/2008, (EU) No 167/2013, (EU) No 168/2013, (EU) 2018/858, (EU) 2018/1139 and (EU) 2019/2144 and Directives 2014/90/EU, (EU) 2016/797 and (EU) 2020/1828 (Artificial Intelligence Act) (OJ L, 12.7.2024, 2024/1689)},
}

@inproceedings{zhou_brainuicl_2024,
 author = {Zhou, Yangxuan and Zhao, Sha and Wang, Jiquan and Jiang, Haiteng and Li, Shijian and Li, Tao and Pan, Gang},
 booktitle = {International Conference on Representation Learning},
 editor = {Y. Yue and A. Garg and N. Peng and F. Sha and R. Yu},
 pages = {89468--89489},
 title = {BrainUICL: An Unsupervised Individual Continual Learning Framework for EEG Applications},
 url = {https://proceedings.iclr.cc/paper_files/paper/2025/file/deb06c4a6d757b51166fef8ab6c607ec-Paper-Conference.pdf},
 volume = {2025},
 year = {2025}
}

@ARTICLE{duan_retain_2024,
  author={Duan, Tiehang and Wang, Zhenyi and Shen, Li and Doretto, Gianfranco and Adjeroh, Donald A. and Li, Fang and Tao, Cui},
  journal={IEEE Transactions on Artificial Intelligence}, 
  title={Retain and Adapt: Online Sequential EEG Classification With Subject Shift}, 
  year={2024},
  volume={5},
  number={9},
  pages={4479-4492},
  doi={10.1109/TAI.2024.3385390}}

@article{duan_online_2024,
title = {Online continual decoding of streaming EEG signal with a balanced and informative memory buffer},
journal = {Neural Networks},
volume = {176},
pages = {106338},
year = {2024},
issn = {0893-6080},
doi = {https://doi.org/10.1016/j.neunet.2024.106338},
url = {https://www.sciencedirect.com/science/article/pii/S0893608024002624},
author = {Tiehang Duan and Zhenyi Wang and Fang Li and Gianfranco Doretto and Donald A. Adjeroh and Yiyi Yin and Cui Tao}
}

@article{aguilar_continuous_2025,
	title = {Continuous metaplastic training on brain signals},
	volume = {2},
	copyright = {2025 The Author(s)},
	issn = {3004-8672},
	url = {https://www.nature.com/articles/s44335-025-00025-5},
	doi = {10.1038/s44335-025-00025-5},
	language = {en},
	number = {1},
	urldate = {2025-05-27},
	journal = {npj Unconventional Computing},
	author = {Aguilar, Isabelle and Bersani--Veroni, Thomas and Herbozo Contreras, Luis Fernando and Love, Thomas and Nikpour, Armin and Querlioz, Damien and Kavehei, Omid},
	month = apr,
	year = {2025},
	note = {Publisher: Nature Publishing Group},
	pages = {1--9},
}

@INPROCEEDINGS{li_personalized_2025,
  author={Li, Dan and Shin, Hye-Bin and Yin, Kang},
  booktitle={2025 13th International Conference on Brain-Computer Interface (BCI)}, 
  title={Personalized Continual EEG Decoding: Retaining and Transferring Knowledge}, 
  year={2025},
  volume={},
  number={},
  pages={1-4},
  url = {http://arxiv.org/abs/2411.11874},
  doi={10.1109/BCI65088.2025.10931666}}

@article{roy_deep_2019,
doi = {10.1088/1741-2552/ab260c},
url = {https://doi.org/10.1088/1741-2552/ab260c},
year = {2019},
month = {aug},
publisher = {IOP Publishing},
volume = {16},
number = {5},
pages = {051001},
author = {Roy, Yannick and Banville, Hubert and Albuquerque, Isabela and Gramfort, Alexandre and Falk, Tiago H and Faubert, Jocelyn},
title = {Deep learning-based electroencephalography analysis: a systematic review},
journal = {Journal of Neural Engineering}
}

@ARTICLE{zhang_mini_2024,
AUTHOR={Zhang, Zhihui  and Fort, Josep M.  and Giménez Mateu, Lluis },
TITLE={Mini review: Challenges in EEG emotion recognition},
JOURNAL={Frontiers in Psychology},
VOLUME={Volume 14 - 2023},
YEAR={2024},
URL={https://www.frontiersin.org/journals/psychology/articles/10.3389/fpsyg.2023.1289816},
DOI={10.3389/fpsyg.2023.1289816},
ISSN={1664-1078}}

@inproceedings{lopez-paz_gradient_2022,
 author = {Lopez-Paz, David and Ranzato, Marc\textquotesingle Aurelio},
 booktitle = {Advances in Neural Information Processing Systems},
 editor = {I. Guyon and U. Von Luxburg and S. Bengio and H. Wallach and R. Fergus and S. Vishwanathan and R. Garnett},
 pages = {},
 publisher = {Curran Associates, Inc.},
 title = {Gradient Episodic Memory for Continual Learning},
 url = {https://proceedings.neurips.cc/paper_files/paper/2017/file/f87522788a2be2d171666752f97ddebb-Paper.pdf},
 volume = {30},
 year = {2017}
}

@article{apicella_toward_2024,
    author = {Apicella, Andrea and Arpaia, Pasquale and D’Errico, Giovanni and Marocco, Davide and Mastrati, Giovanna and Moccaldi, Nicola and Prevete, Roberto},
    title = {Toward cross-subject and cross-session generalization in EEG-based emotion recognition: Systematic review, taxonomy, and methods},
    year = {2024},
    issue_date = {Nov 2024},
    publisher = {Elsevier Science Publishers B. V.},
    address = {NLD},
    volume = {604},
    number = {C},
    issn = {0925-2312},
    url = {https://doi.org/10.1016/j.neucom.2024.128354},
    doi = {10.1016/j.neucom.2024.128354},
    journal = {Neurocomput.},
    month = nov,
    numpages = {23}
}

@Article{chaddad_electroencephalography_2023,
    AUTHOR = {Chaddad, Ahmad and Wu, Yihang and Kateb, Reem and Bouridane, Ahmed},
    TITLE = {Electroencephalography Signal Processing: A Comprehensive Review and Analysis of Methods and Techniques},
    JOURNAL = {Sensors},
    VOLUME = {23},
    YEAR = {2023},
    NUMBER = {14},
    ARTICLE-NUMBER = {6434},
    URL = {https://www.mdpi.com/1424-8220/23/14/6434},
    PubMedID = {37514728},
    ISSN = {1424-8220},
    DOI = {10.3390/s23146434}
}

@ARTICLE{saha_intra-_2020,
AUTHOR={Saha, Simanto  and Baumert, Mathias },
TITLE={Intra- and Inter-subject Variability in EEG-Based Sensorimotor Brain Computer Interface: A Review},
JOURNAL={Frontiers in Computational Neuroscience},
VOLUME={Volume 13 - 2019},
YEAR={2020},
URL={https://www.frontiersin.org/journals/computational-neuroscience/articles/10.3389/fncom.2019.00087},
DOI={10.3389/fncom.2019.00087},
ISSN={1662-5188}}

@article{Smith6136,
	author = {Smith, Stephen and Duff, Eugene and Groves, Adrian and Nichols, Thomas E. and Jbabdi, Saad and Westlye, Lars T. and Tamnes, Christian K. and Engvig, Andreas and Walhovd, Kristine B. and Fjell, Anders M. and Johansen-Berg, Heidi and Douaud, Gwena{\"e}lle},
	title = {Structural Variability in the Human Brain Reflects Fine-Grained Functional Architecture at the Population Level},
	volume = {39},
	number = {31},
	pages = {6136--6149},
	year = {2019},
	doi = {10.1523/JNEUROSCI.2912-18.2019},
	publisher = {Society for Neuroscience},
	issn = {0270-6474},
	URL = {https://www.jneurosci.org/content/39/31/6136},
	eprint = {https://www.jneurosci.org/content/39/31/6136.full.pdf},
	journal = {Journal of Neuroscience}
}

@article{bazargani_emotion_2023,
	title = {An {Emotion} {Recognition} {Embedded} {System} using a {Lightweight} {Deep} {Learning} {Model}},
	volume = {13},
	issn = {2228-7477},
	url = {https://journals.lww.com/10.4103/jmss.jmss_59_22},
	doi = {10.4103/jmss.jmss_59_22},
	number = {4},
	journal = {Journal of Medical Signals \& Sensors},
	author = {Bazargani, Mehdi and Tahmasebi, Amir and Yazdchi, Mohammadreza and Baharlouei, Zahra},
	month = oct,
	year = {2023},
	pages = {272--279},
}

@article{shallowconvnet,
author = {Schirrmeister, Robin Tibor and Springenberg, Jost Tobias and Fiederer, Lukas Dominique Josef and Glasstetter, Martin and Eggensperger, Katharina and Tangermann, Michael and Hutter, Frank and Burgard, Wolfram and Ball, Tonio},
title = {Deep learning with convolutional neural networks for EEG decoding and visualization},
journal = {Human Brain Mapping},
volume = {38},
number = {11},
pages = {5391-5420},
doi = {https://doi.org/10.1002/hbm.23730},
url = {https://onlinelibrary.wiley.com/doi/abs/10.1002/hbm.23730},
eprint = {https://onlinelibrary.wiley.com/doi/pdf/10.1002/hbm.23730},
year = {2017}
}

@article{krusienski_critical_2011,
doi = {10.1088/1741-2560/8/2/025002},
url = {https://doi.org/10.1088/1741-2560/8/2/025002},
year = {2011},
month = {mar},
publisher = {},
volume = {8},
number = {2},
pages = {025002},
author = {Krusienski, Dean J and Grosse-Wentrup, Moritz and Galán, Ferran and Coyle, Damien and Miller, Kai J and Forney, Elliott and Anderson, Charles W},
title = {Critical issues in state-of-the-art brain–computer interface signal processing},
journal = {Journal of Neural Engineering}
}

@inproceedings{aljundi_memory_2018,
author = {Aljundi, Rahaf and Babiloni, Francesca and Elhoseiny, Mohamed and Rohrbach, Marcus and Tuytelaars, Tinne},
title = {Memory Aware Synapses: Learning What (not) to Forget},
year = {2018},
isbn = {978-3-030-01218-2},
publisher = {Springer-Verlag},
address = {Berlin, Heidelberg},
url = {https://doi.org/10.1007/978-3-030-01219-9_9},
doi = {10.1007/978-3-030-01219-9_9},
booktitle = {Computer Vision – ECCV 2018: 15th European Conference, Munich, Germany, September 8–14, 2018, Proceedings, Part III},
pages = {144–161},
numpages = {18},
location = {Munich, Germany}
}

@inproceedings{zhao_statistical_2024,
author = {Zhao, Xuyang and Wang, Huiyuan and Huang, Weiran and Lin, Wei},
title = {A statistical theory of regularization-based continual learning},
year = {2024},
publisher = {JMLR.org},
url = {http://arxiv.org/abs/2406.06213},
booktitle = {Proceedings of the 41st International Conference on Machine Learning},
articleno = {2523},
numpages = {19},
location = {Vienna, Austria},
series = {ICML'24}
}

@InProceedings{benzing_unifying_2022,
  title = 	 { Unifying Importance Based Regularisation Methods for Continual Learning },
  author =       {Benzing, Frederik},
  booktitle = 	 {Proceedings of The 25th International Conference on Artificial Intelligence and Statistics},
  pages = 	 {2372--2396},
  year = 	 {2022},
  editor = 	 {Camps-Valls, Gustau and Ruiz, Francisco J. R. and Valera, Isabel},
  volume = 	 {151},
  series = 	 {Proceedings of Machine Learning Research},
  month = 	 {28--30 Mar},
  publisher =    {PMLR},
  url = 	 {https://proceedings.mlr.press/v151/benzing22a.html}
}

@INPROCEEDINGS{kann_evaluation_2023,
  author={Kann, Bonpagna and Castellanos-Paez, Sandra and Lalanda, Philippe},
  booktitle={2023 IEEE International Conference on Pervasive Computing and Communications Workshops and other Affiliated Events (PerCom Workshops)}, 
  title={Evaluation of Regularization-based Continual Learning Approaches: Application to HAR}, 
  year={2023},
  volume={},
  number={},
  pages={460-465},
  doi={10.1109/PerComWorkshops56833.2023.10150281}}

@InProceedings{pmlr-v108-barshan20a,
  title = 	 {RelatIF: Identifying Explanatory Training Samples via Relative Influence},
  author =       {Barshan, Elnaz and Brunet, Marc-Etienne and Dziugaite, Gintare Karolina},
  booktitle = 	 {Proceedings of the Twenty Third International Conference on Artificial Intelligence and Statistics},
  pages = 	 {1899--1909},
  year = 	 {2020},
  editor = 	 {Chiappa, Silvia and Calandra, Roberto},
  volume = 	 {108},
  series = 	 {Proceedings of Machine Learning Research},
  month = 	 {26--28 Aug},
  publisher =    {PMLR},
  url = 	 {https://proceedings.mlr.press/v108/barshan20a.html}
}

@misc{pascanu_revisiting_2014,
	title = {Revisiting {Natural} {Gradient} for {Deep} {Networks}},
	url = {http://arxiv.org/abs/1301.3584},
	doi = {10.48550/arXiv.1301.3584},
	urldate = {2025-08-03},
	publisher = {arXiv},
	author = {Pascanu, Razvan and Bengio, Yoshua},
	month = feb,
	year = {2014},
	note = {arXiv:1301.3584 [cs]},
}

@INPROCEEDINGS{meshkov_convnets_2024,
  author={Meshkov, Vladislav and Kiselev, Nikita and Grabovoy, Andrey},
  booktitle={2024 Ivannikov Ispras Open Conference (ISPRAS)}, 
  title={ConvNets Landscape Convergence: Hessian-Based Analysis of Matricized Networks}, 
  year={2024},
  volume={},
  number={},
  pages={1-10},
  url = {https://ieeexplore.ieee.org/document/10899113},
  doi={10.1109/ISPRAS64596.2024.10899113}}

@inproceedings{kunstner_limitations_2019,
 author = {Kunstner, Frederik and Hennig, Philipp and Balles, Lukas},
 booktitle = {Advances in Neural Information Processing Systems},
 editor = {H. Wallach and H. Larochelle and A. Beygelzimer and F. d\textquotesingle Alch\'{e}-Buc and E. Fox and R. Garnett},
 pages = {},
 publisher = {Curran Associates, Inc.},
 title = {Limitations of the empirical Fisher approximation for natural gradient descent},
 url = {https://proceedings.neurips.cc/paper_files/paper/2019/file/46a558d97954d0692411c861cf78ef79-Paper.pdf},
 volume = {32},
 year = {2019}
}

@misc{ven_computation_2025,
	title = {On the {Computation} of the {Fisher} {Information} in {Continual} {Learning}},
	url = {http://arxiv.org/abs/2502.11756},
	doi = {10.48550/arXiv.2502.11756},
	urldate = {2025-08-03},
	publisher = {arXiv},
	author = {Ven, Gido M. van de},
	month = feb,
	year = {2025},
	note = {arXiv:2502.11756 [cs]},
}

@article{cho_neuromimetic_2025,
title = {Neuromimetic metaplasticity for adaptive continual learning without catastrophic forgetting},
journal = {Neural Networks},
volume = {190},
pages = {107762},
year = {2025},
issn = {0893-6080},
doi = {https://doi.org/10.1016/j.neunet.2025.107762},
url = {https://www.sciencedirect.com/science/article/pii/S0893608025006422},
author = {Suhee Cho and Hyeonsu Lee and Seungdae Baek and Se-Bum Paik}
}

@misc{chaudhry_efficient_2019,
	title={Efficient Lifelong Learning with A-GEM},
    author={Chaudhry, Arslan and Ranzato, Marc’Aurelio and Rohrbach, Marcus and Elhoseiny, Mohamed},
    booktitle={ICLR},
	url = {http://arxiv.org/abs/1812.00420},
	doi = {10.48550/arXiv.1812.00420},
	urldate = {2025-05-17},
	month = jan,
	year = {2019},
}

@article{cognemosense,
	doi = {10.20944/preprints202505.0072.v1},
	url = {https://doi.org/10.20944/preprints202505.0072.v1},
	year = 2025,
	month = {May},
	publisher = {Preprints},
	author = {Hakan Dalkıç},
	title = {CognEmoSense: A Continual Learning and Context-Aware EEG Emotion Recognition System Using Transformer-Augmented Brain-State Modeling},
	journal = {Preprints}
}

@InProceedings{schwarz_progress_2018,
  title = 	 {Progress \& Compress: A scalable framework for continual learning},
  author =       {Schwarz, Jonathan and Czarnecki, Wojciech and Luketina, Jelena and Grabska-Barwinska, Agnieszka and Teh, Yee Whye and Pascanu, Razvan and Hadsell, Raia},
  booktitle = 	 {Proceedings of the 35th International Conference on Machine Learning},
  pages = 	 {4528--4537},
  year = 	 {2018},
  editor = 	 {Dy, Jennifer and Krause, Andreas},
  volume = 	 {80},
  series = 	 {Proceedings of Machine Learning Research},
  month = 	 {10--15 Jul},
  publisher =    {PMLR},
  url = 	 {https://proceedings.mlr.press/v80/schwarz18a.html}
}

\clearpage
\appendix
\section{Appendix}
\subsection{Figures}

\subsubsection{Preliminary Experiment}

\begin{center}
    \begin{minipage}{\linewidth}
        \centering
        
        \includegraphics[width=\linewidth]{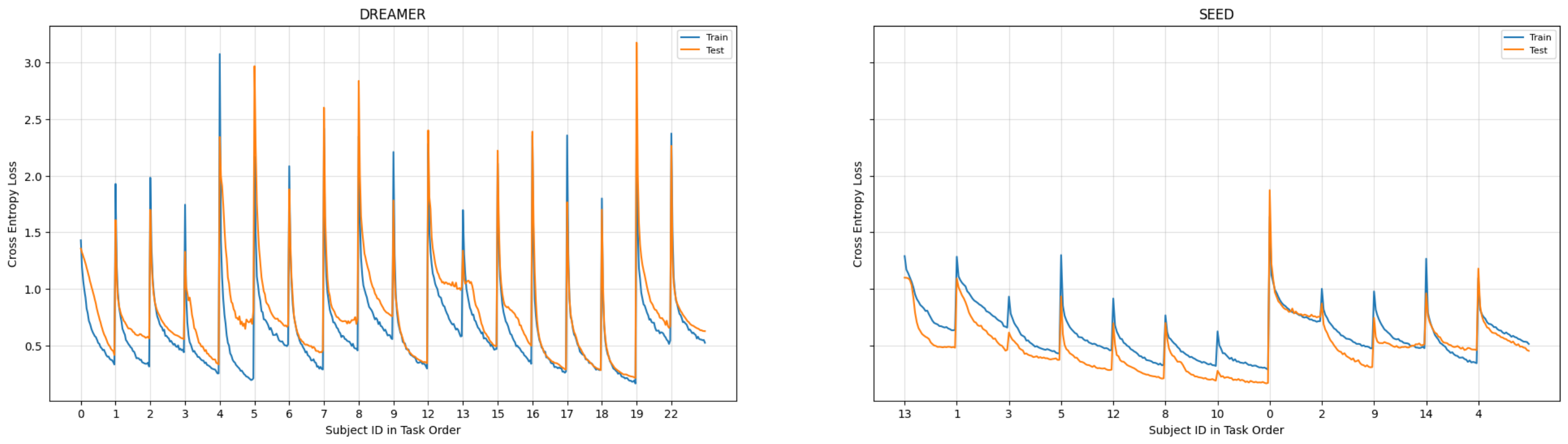}
        \captionof{figure}{[Naïve] Train-Test Loss curves for both DREAMER and SEED, where loss is a measure of Cross Entropy Loss.}
        \label{fig:train-test-loss}

        \vspace{5mm}
    
        \includegraphics[width=\linewidth]{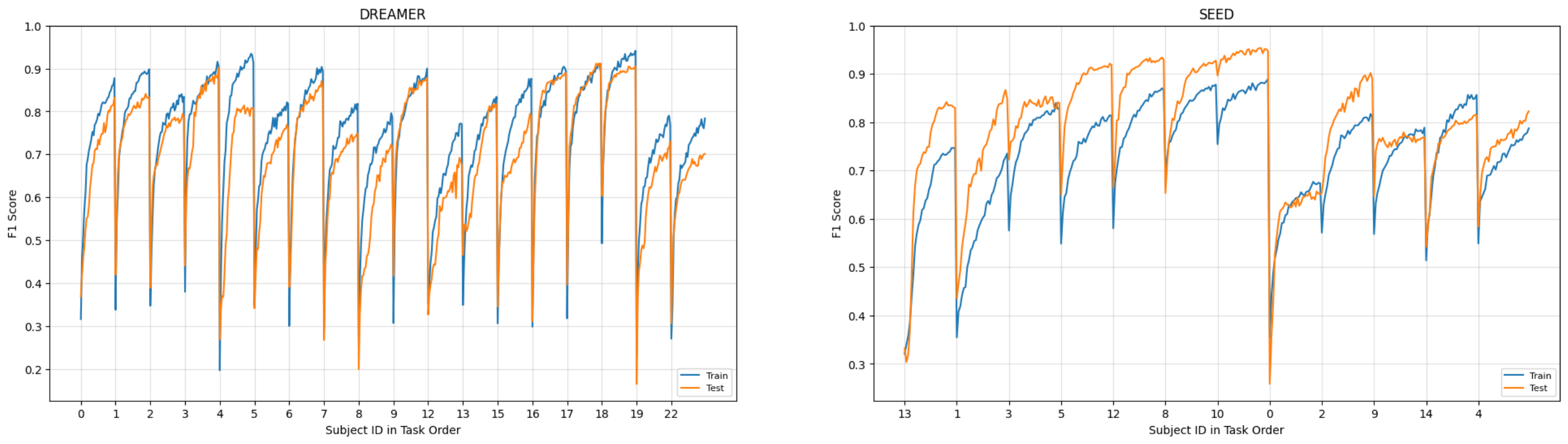}
        \captionof{figure}{[Naïve] Train-Test F1 score curves for both DREAMER and SEED.}
        \label{fig:train-test-accuracy}

        \vspace{5mm}
    
        \includegraphics[width=\linewidth]{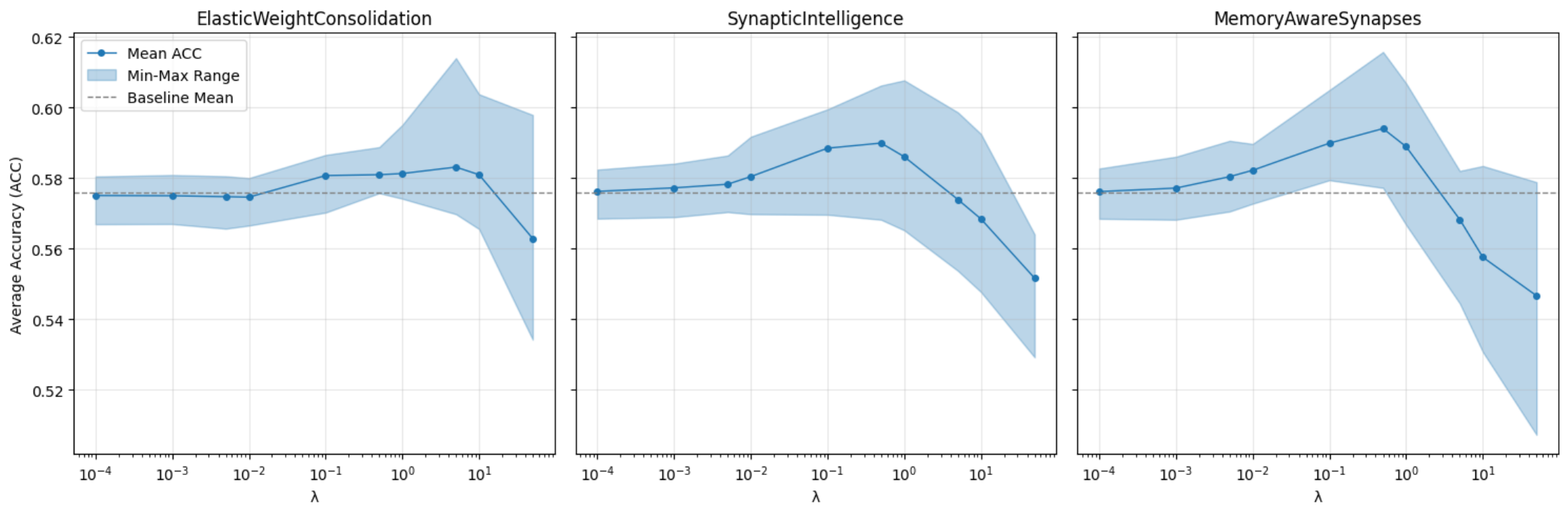}
        \captionof{figure}{[SEED] Average ACC across five seeds for each strategy with respect to increased regularisation ($\lambda$) upon repeated training.}
        \label{fig:seed_acc}
    \end{minipage}
\end{center}

\begin{minipage}{\linewidth}
    \centering

    \includegraphics[width=\linewidth]{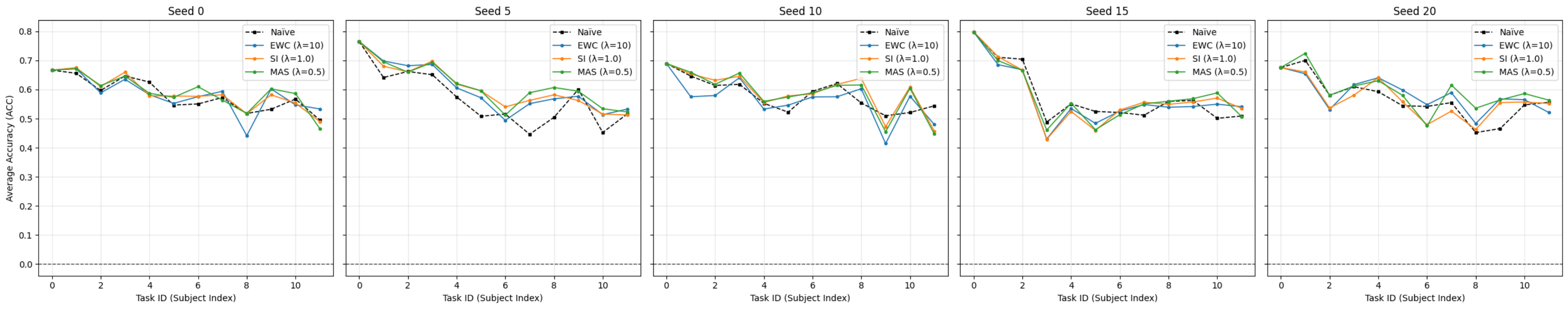}
    \captionof{figure}{[SEED] ACC across tasks for each strategy and each seed, where EWC $\lambda = 10.0$, SI $\lambda = 1.0$, and MAS $\lambda = 0.5$.}
    \label{fig:seed_acc_lambda1}

    \vspace{5mm}

    \includegraphics[width=\linewidth]{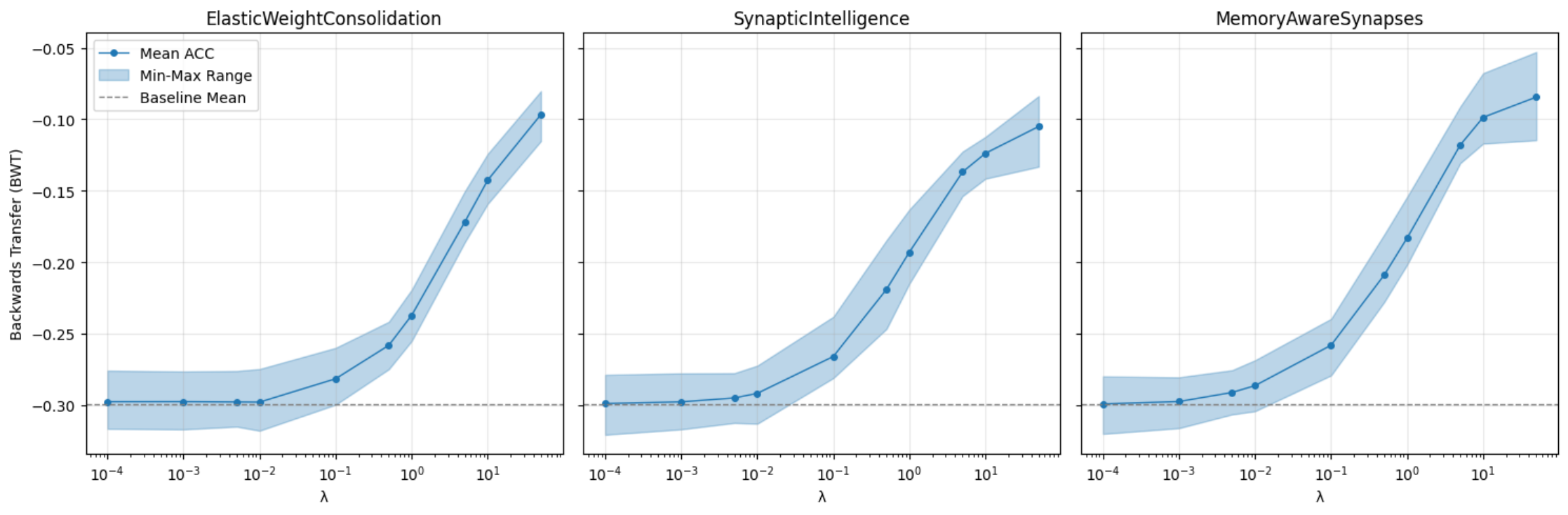}
    \captionof{figure}{[SEED] Average BWT across seeds for each strategy with respect to increased regularisation ($\lambda$).}
    \label{fig:seed_bwt}

    \vspace{5mm}

\end{minipage}

\begin{figure}[hbt!]
    \centering
    \includegraphics[width=\linewidth]{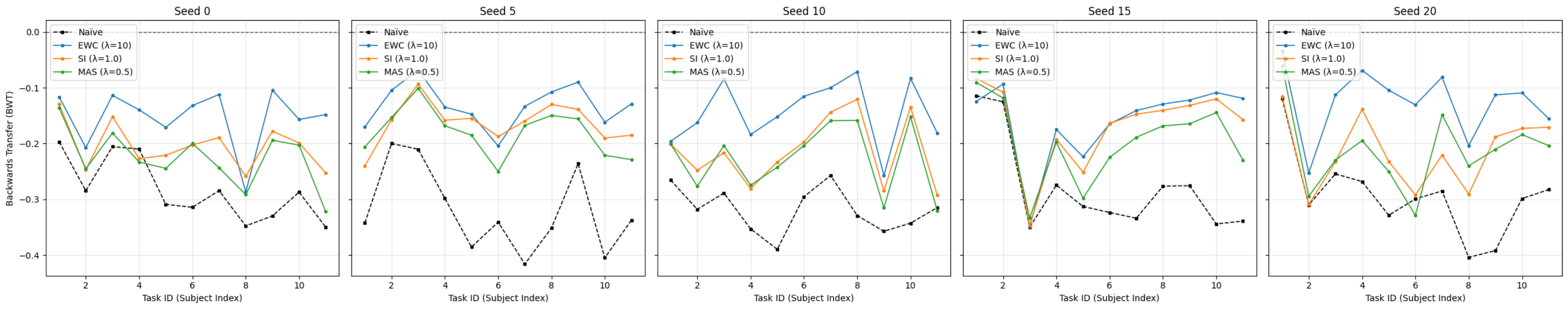}
    \caption{[SEED] BWT across tasks for each strategy and each seed, where EWC $\lambda = 10.0$, SI $\lambda = 1.0$, and MAS $\lambda = 0.5$.}
    \label{fig:seed_bwt_lambda1}
\end{figure}

\vspace{5mm}

\begin{figure}[hbt!]
    \centering
    \includegraphics[width=\linewidth]{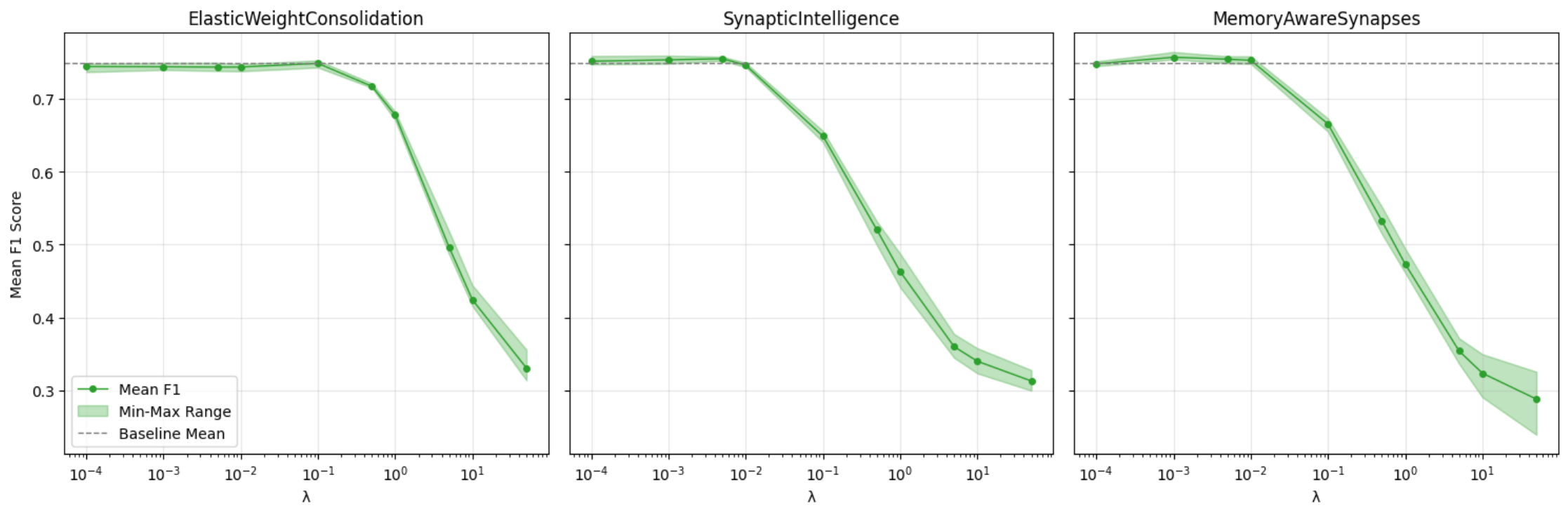}
    \caption{[DREAMER] Average F1 score across seeds for each strategy with respect to increased regularisation ($\lambda$).}
    \label{fig:dreamer_f1}
\end{figure}

\begin{figure}[hbt!]
    \centering
    \includegraphics[width=\linewidth]{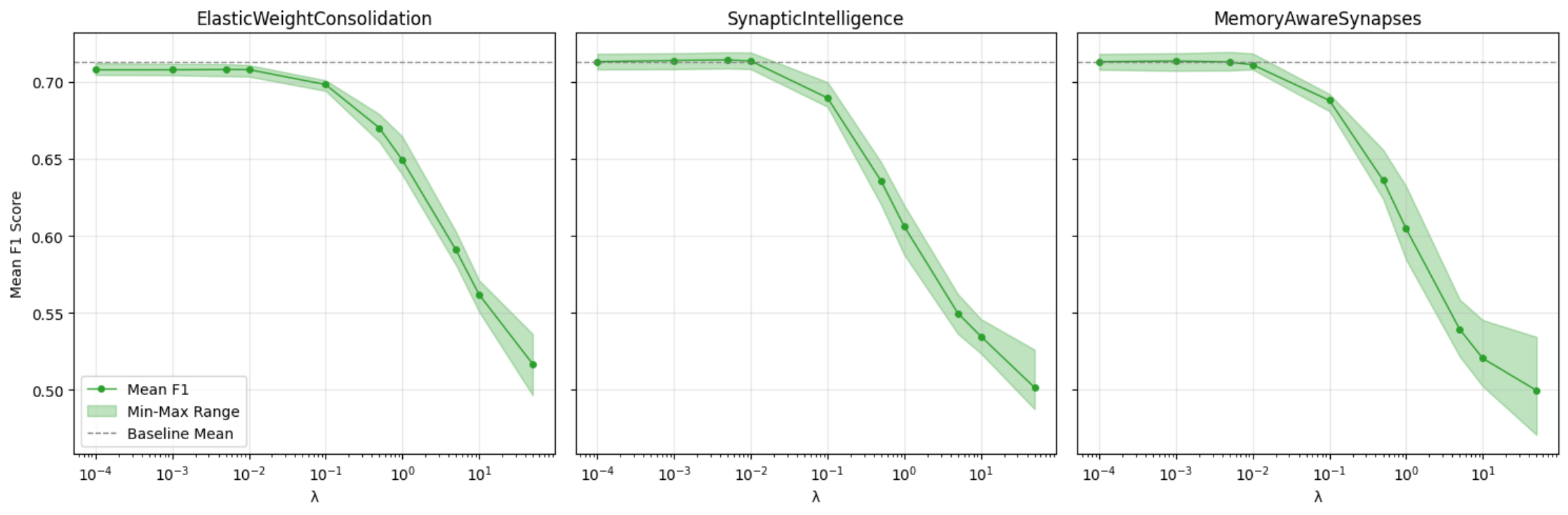}
    \caption{[SEED] Average F1 score across seeds for each strategy with respect to increased regularisation ($\lambda$).}
    \label{fig:seed_f1}
\end{figure}

\vspace{5cm}

\subsubsection{H2: Subsequent subjects require conflicting parameter updates, leading to gradient interference rather than convergence, limiting generalisability.}

\label{sec:h2-appendix}

\begin{figure}[ht!]
    \centering
    \includegraphics[width=\linewidth]{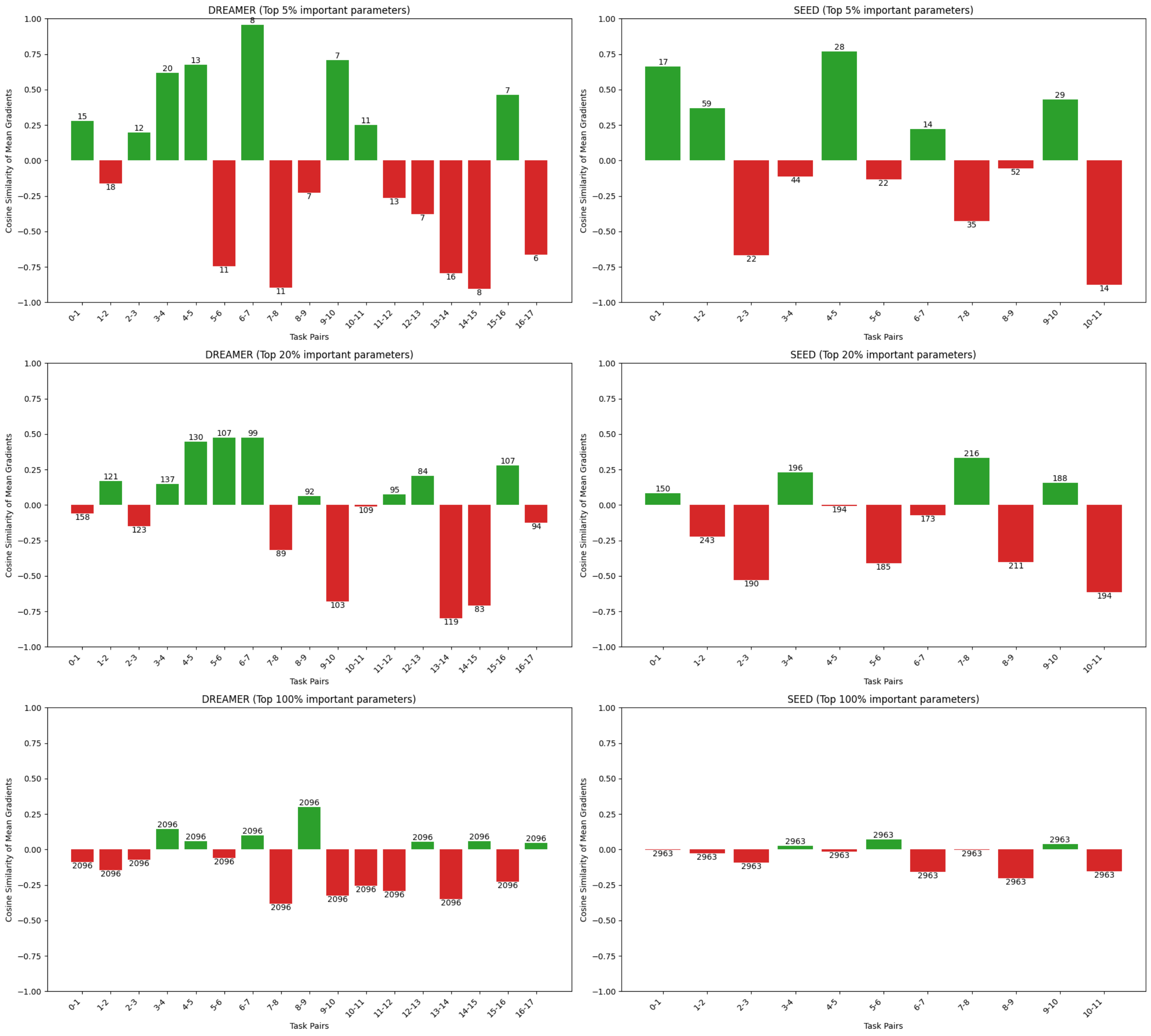}
    \caption{[SI] Cosine similarity of gradients of parameters in top 5\%, 20\%, 100\% of both tasks in task pair. The label above or below each bar indicates the number of parameters that were in this overlap.}
    \label{fig:si_cosine}
\end{figure}

\begin{figure}[hbt!]
    \centering
    \includegraphics[width=\linewidth]{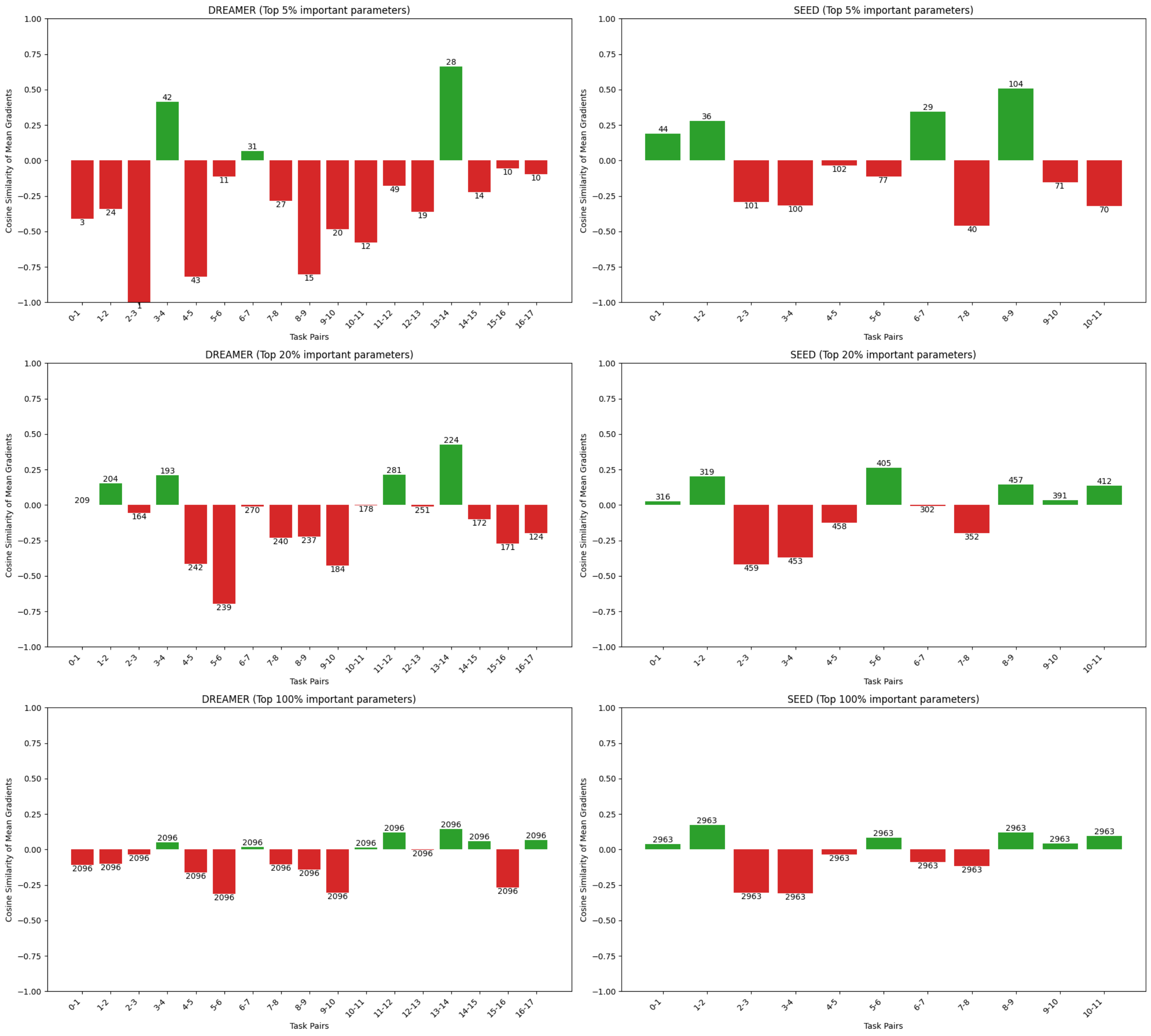}
    \caption{[MAS] Cosine similarity of gradients of parameters in top 5\%, 20\%, 100\% of both tasks in task pair. The label above or below each bar indicates the number of parameters that were in this overlap.}
    \label{fig:mas_cosine}
\end{figure}

\clearpage
\subsubsection{H3: Importance accumulates across tasks, constraining plasticity in the network early, limiting scalability.}
\vspace{5mm}

\begin{center}
    
\includegraphics[width=0.9\linewidth]{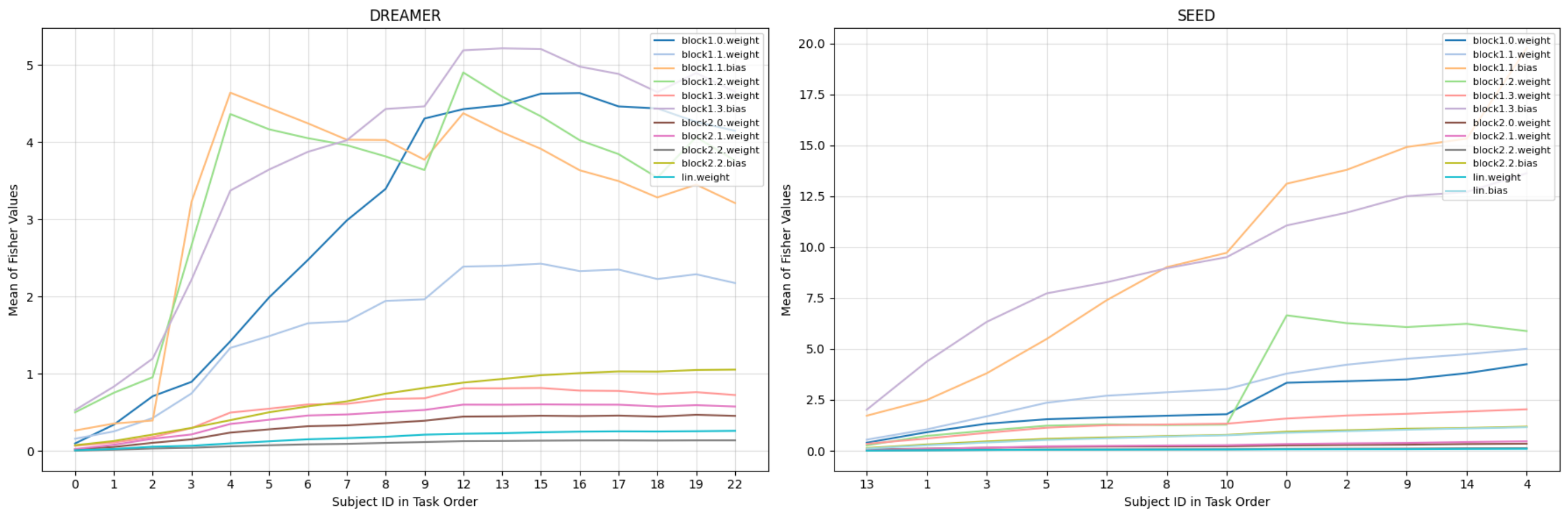}
\captionof{figure}{[EWC] Mean importance per named parameter across tasks, where importance is a measure of Fisher Information.}
\label{fig:ewc_h2_omegacurve}

\vspace{5mm}
\includegraphics[width=0.9\linewidth]{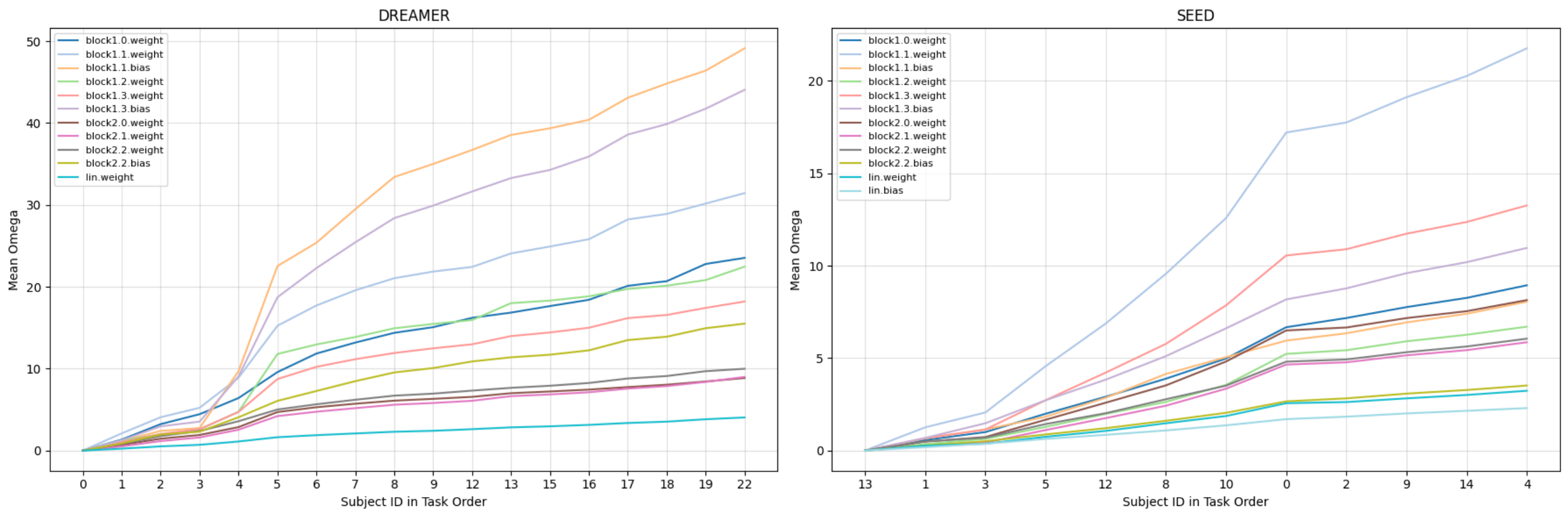}
\captionof{figure}{[MAS] Mean importance per named parameter across tasks, where importance is a measure of sensitivity of the network output to changes in each parameter.}
\label{fig:mas_h2_omegacurve}

\vspace{5mm}
\includegraphics[width=0.9\linewidth]{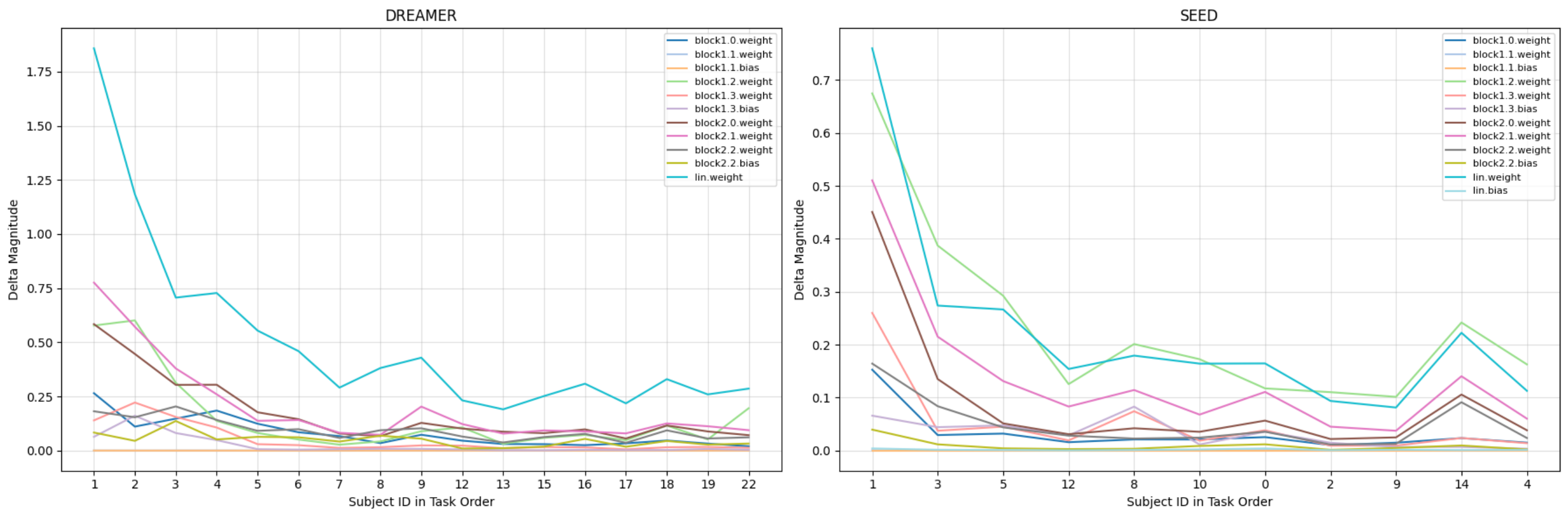}
\captionof{figure}{[EWC] Magnitude of parameter changes (L2 norm) across tasks, per named parameter.}
\label{fig:ewc_h2_meandelta}

\vspace{5mm}
\includegraphics[width=0.9\linewidth]{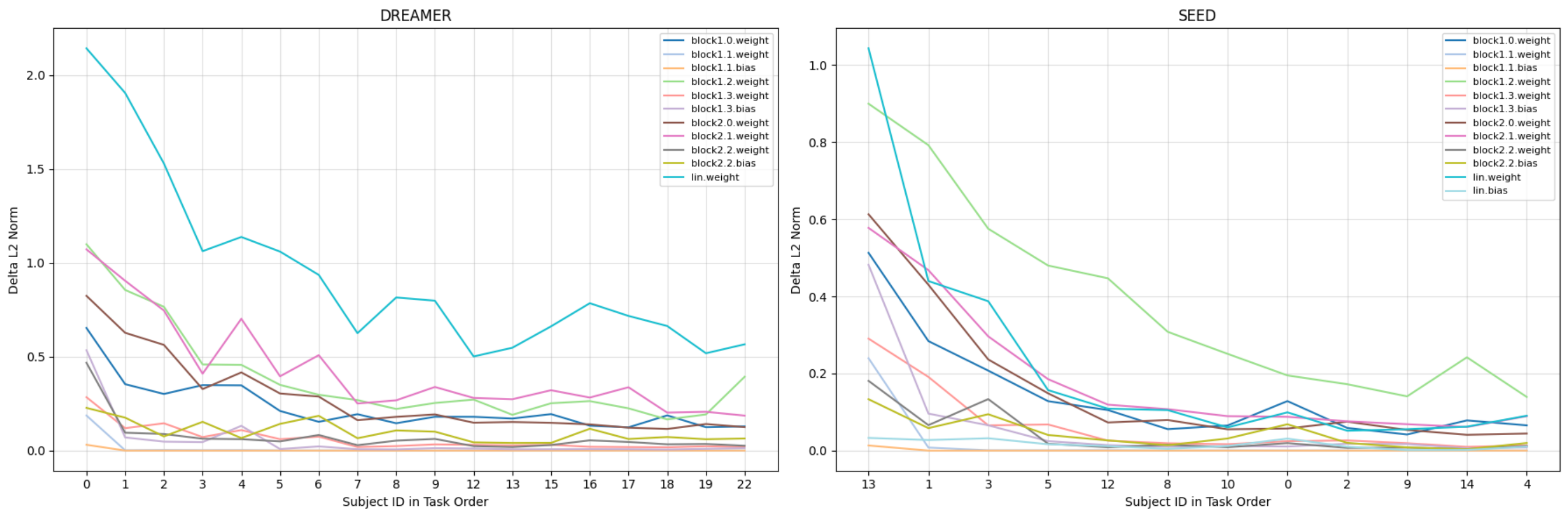}
\captionof{figure}{[MAS] Magnitude of parameter changes (L2 norm) across tasks, per named parameter.}
\label{fig:mas_h2_meandelta}

\end{center}

\begin{figure}[hbt!]
    \centering
    \captionsetup{width=0.9\linewidth}
    \includegraphics[width=0.9\linewidth]{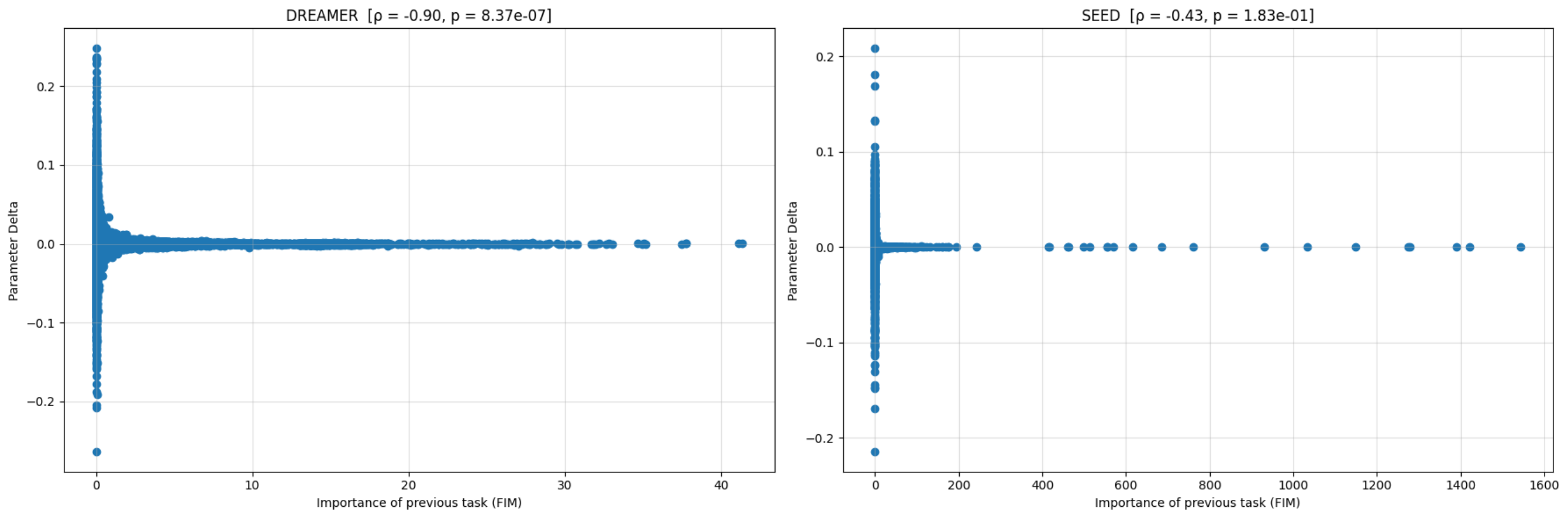}
    \caption{[EWC] Correlation between importance of the previous task and the parameter change in the subsequent task.}
    \label{fig:ewc_h2_deltascatter}
\end{figure}

\begin{figure}[hbt!]
    \centering
    \captionsetup{width=0.9\linewidth}
    \includegraphics[width=0.9\linewidth]{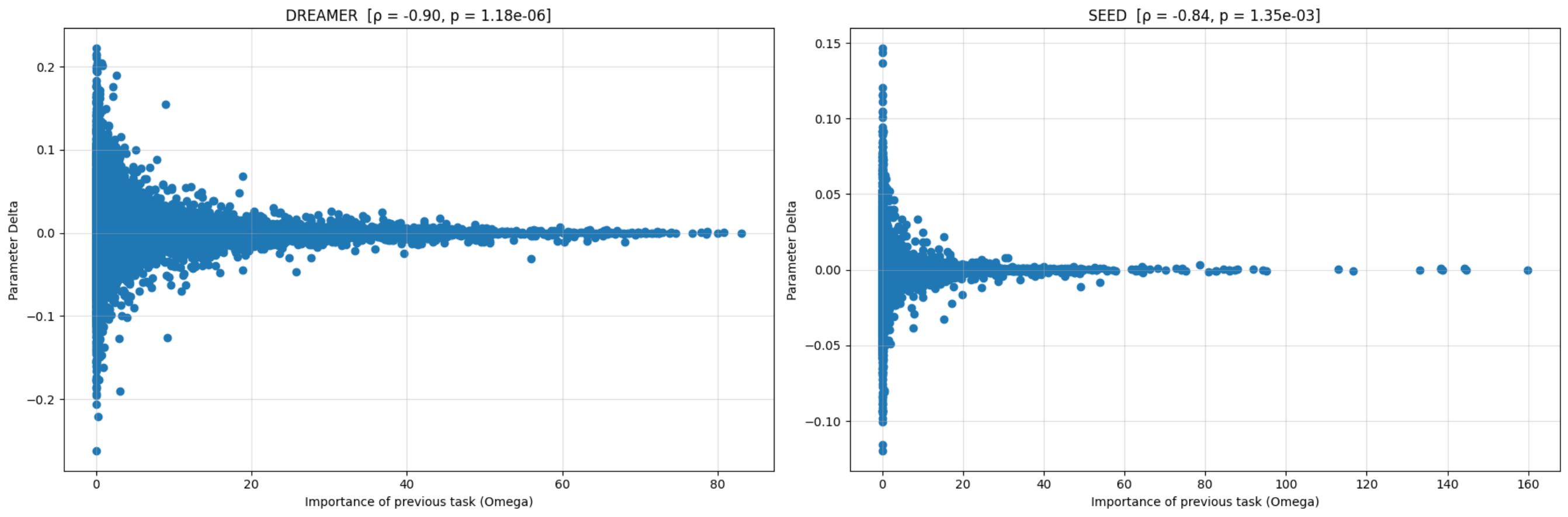}
    \caption{[MAS] Correlation between importance of the previous task and the parameter change in the subsequent task.}
    \label{fig:mas_h2_deltascatter}
\end{figure}

\subsubsection{H4: Subject order impacts which parameters are constrained early, making each method highly sensitive to random seed.}
\label{sec:appendix-h4}
\begin{figure}[hbt!]
    \centering
    \includegraphics[width=0.65\linewidth]{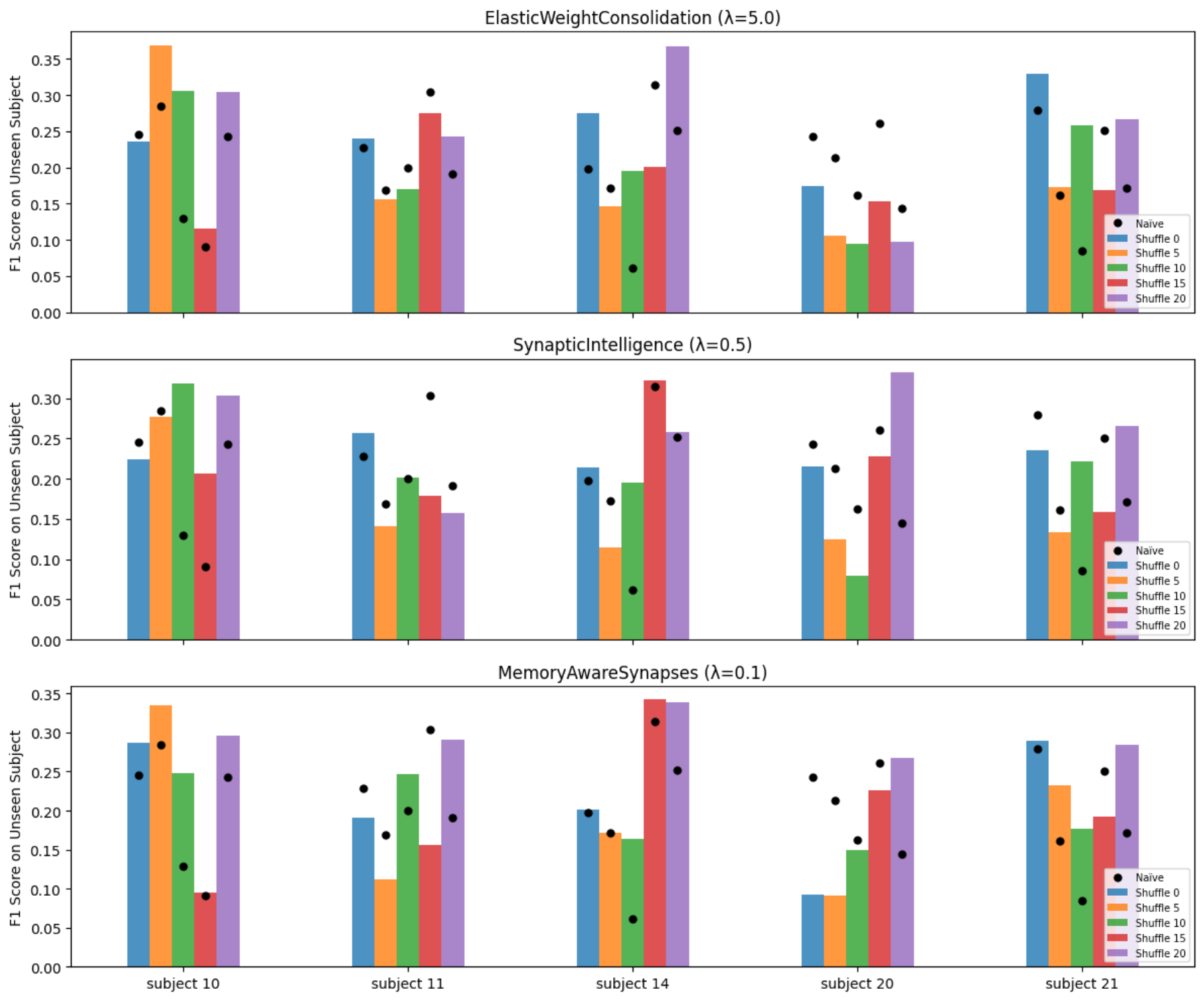}
    \caption{[DREAMER] Visualisation of the lack of generalisability of each strategy on the DREAMER dataset, with respect to both random seed and the baseline. The black dots indicate baseline performance, whereas each colour represents a different subject-order shuffle.}
    \label{fig:dreamer_shuffle}
\end{figure}
\begin{figure}[hbt!]
    \centering
    \includegraphics[width=0.65\linewidth]{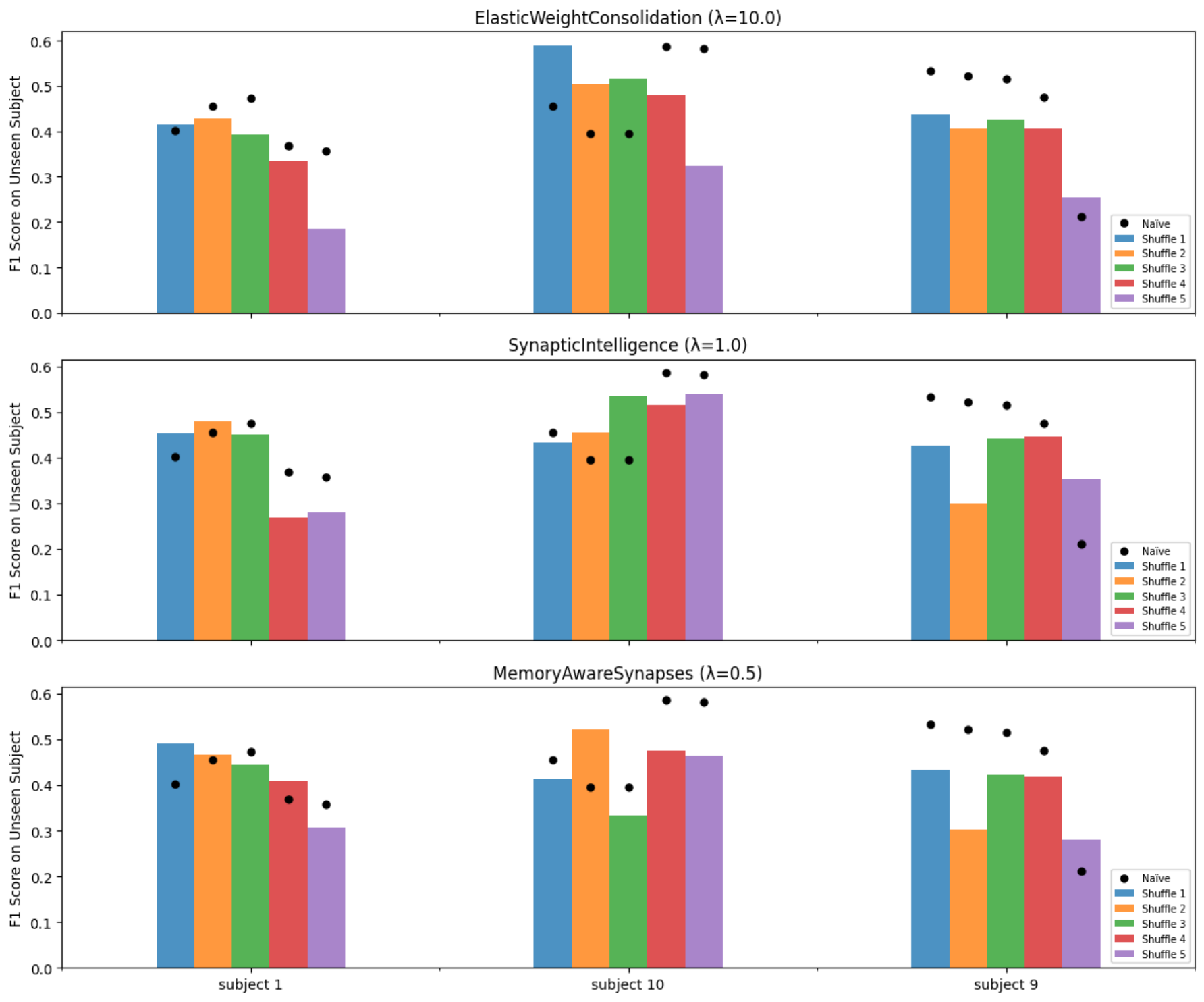}
    \caption{[SEED] Visualisation of the lack of generalisability of each strategy on the SEED dataset, with respect to both random seed and the baseline. The black dots indicate baseline performance, whereas each colour represents a different subject-order shuffle.}
    \label{fig:seed_shuffle}
\end{figure}

\clearpage

\subsection{Mathematics}

\subsubsection{Proof that the Fisher Information equals the expected Hessian matrix of the NLL}
\label{sec:fisher-hessian}
As identified in \cite{pmlr-v108-barshan20a, kirkpatrick_overcoming_2017, pascanu_revisiting_2014, kunstner_limitations_2019}, the following holds under regularity conditions.
\allowdisplaybreaks
{\tiny\begin{align} 
    F(\theta) &= \frac{1}{n}\sum^n_{i=1}\mathbb{E}_{p_\theta(y|x_i)}\left[\nabla_\theta \log p_\theta (y|x_i)\nabla_\theta \log p_\theta (y|x_i)^\top\right] \\
    &= \frac{1}{n}\sum^n_{i=1}\left(\underbrace{\mathbb{E}_{p_\theta(y|x_i)}\left[\frac{\nabla^2_\theta p_\theta(y|x_i)}{p_\theta(y|x_i)}\right]}_{= 0\text{ under regularity conditions}}-\mathbb{E}_{p_\theta(y|x_i)}\left[\nabla^2_\theta\log p_\theta (y|x_i)\right]\right) \\
    &= \frac{1}{n}\sum^n_{i=1}\mathbb{E}_{p_\theta(y|x_i)}\left[-\nabla^2_\theta\log p_\theta (y|x_i)\right]\\
    &\approx -\frac{1}{n}\sum^n_{i=1} \nabla^2_\theta\log p_\theta (y|x_i), \quad\text{under the assumption }p_\theta\approx p_\text{data}\\
    & = H(\theta)
\end{align}}

\subsubsection{Impact of stochastic batch noise and intra-subject variability of Adam on Synaptic Intelligence}
\label{sec:path-integral-adam-math}
With Adam:
\begin{equation}
    \theta' \leftarrow \theta - \eta_t\cdot\frac{m_t}{\sqrt{v_t} + \xi},
\end{equation}
where
\begin{align}
    m_t &\leftarrow \beta_1 m_{t-1} + (1-\beta_1) g_t\\
    v_t &\leftarrow \beta_2 v_{t-1} + (1-\beta_2) g_t^2.
\end{align}
This unrolls to
\begin{align}
    m_t &=(1-\beta_1)\sum^t_{i=1}\beta^{t-i}_1 g_i\\
    v_t &=(1-\beta_2)\sum^t_{i=1}\beta^{t-i}_2 g_i^2.
\end{align}
The path integral then becomes
\begin{align}
    \omega^\nu_k&\approx-\sum_{t\leqslant T} g_t\left(-\eta_t\cdot\frac{m_t}{\sqrt{v_t}+\xi}\right)
    =\sum_{t\leqslant T}\frac{\eta_tg_k(t)m_t}{\sqrt{v_t}+\xi}.\\
\end{align}
When gradients are noisy, $v_t$ becomes large as we are squaring the same gradient. $\sum_tg_k(t)m_t\rightarrow (1-\beta_1)g^2_t$. Benzing \cite{benzing_unifying_2022} describes this assumption as $(1-\beta_1)\sigma^2_t\gg\beta_1m_{t-1}g_t$.

Consider again, $g'_t = (g_t + \sigma_t + \mu_t)$, as well as the definition of first and second moment of the gradient as used in Adam, this gives the following (see Appendix 
for step by step derivation):
\begin{align}
\label{eq:si_decomp}
    &\sum_{t\leqslant T}\frac{\eta_tg_tm_t}{\sqrt{v_t}+\xi} \\
    &= \sum_{t\leqslant T}\frac{\eta_t (1-\beta_1)(g_t + \sigma_t + \mu_t)^2 + \eta_t\beta_1(g_t + \sigma_t + \mu_t)m_{t-1}}{\sqrt{v_t}+\xi} \\
     &\approx \sum_{t\leqslant T}\frac{\eta_t(1-\beta_1)(g_t + \sigma_t + \mu_t)^2}{\sqrt{v_t} + \xi}+\frac{\epsilon}{\sqrt{v_T}}
\end{align}
under \cite{benzing_unifying_2022}'s assumption that gradient noise is larger than the gradient, or that as the gradients oscillate over many timesteps, $\sum_{t\leqslant T} \frac{\eta_t\beta_1g_tm_{t-1}}{\sqrt{v_t} + \xi}\rightarrow 0$. This shows that with Adam too, the path integral depends on the accumulation of noise across time. This means that under intra-subject variability and stochastic noise, the path integral is not a stable measure of importance but instead depends on $T$, when training stopped.
Benzing \cite{benzing_unifying_2022} further investigates this by claiming per Equation \eqref{eq:benzing} that the path integral is proportional to the square root of the Hessian, echoeing Section~\ref{sec:fisher} where it was stated that the Fisher Information is a measure of noise sensitiviy rather than curvature when $p_\theta\not\approx p_{\text{data}}$.
\begin{align}
\label{eq:benzing}
    \omega^\nu&\approx\sum_{t\leqslant T}\frac{\eta_t(1-\beta_1)(g_t + \sigma_t + \mu_t)^2}{\sqrt{v_t} + \xi}\\
    &=\frac{(1-\beta_1)}{\sqrt{v_T}}\sum_{t\leqslant T}\eta_t\sqrt{\frac{v_T}{v_t}}(g_t + \sigma_t + \mu_t)^2\\
    &\propto \frac{1-\beta_1}{\sqrt{v_T}} v_T\\
    &\propto \sqrt{v_T}
\end{align}
According to Benzing, given $v_T$ is a common approximation of the Fisher Information, our previous discussion on mismatch between the Fisher Information and the Hessian matrix holds. 

This aligns with Benzing's investigation \cite{benzing_unifying_2022}, which demonstrates that the path integral is essentially proportional to the square root of the final moment term:
\begin{align}
\label{eq:benzing}
    \omega^\nu&\approx\sum_{t\leqslant T}\frac{\eta_t(1-\beta_1)(g_t + \sigma_t + \mu_t)^2}{\sqrt{v_t} + \xi}\\
    &=\frac{(1-\beta_1)}{\sqrt{v_T} + \xi}\sum_{t\leqslant T}\eta_t\sqrt{\frac{v_T}{v_t}}(g_t + \sigma_t + \mu_t)^2\\
    &\propto \frac{1-\beta_1}{\sqrt{v_T}+ \xi} v_T\\
    &\propto \sqrt{v_T}.
\end{align}
Given that $v_T$ is a common approximation for the diagonal of the empirical Fisher Information matrix, this echoes our theoretical evaluation of EWC. SI's importance heuristic is fundamentally proportional to the empirical Fisher Information \cite{benzing_unifying_2022}, a heuristic we have established is likely more sensitive to noise than true curvature when applied to EEG signals. It is evident from \eqref{eq:si_decomp}, that the path integral implicitly approximates the squared gradients too. If $\omega^\nu_k$ is not representative of true parameter importance, neither is $\Omega_k$. While Adam's momentum based mechanism dampens the impact of gradient fluctuations compared to vanilla SGD, the underlying problem of unreliable importance estimates remains.

\end{document}